\documentclass[conference]{IEEEtran}
\usepackage{times}
\usepackage{amssymb}
\usepackage{amsmath}
\usepackage[numbers]{natbib}
\usepackage{multicol}
\usepackage[bookmarks=true,bookmarksdepth=3]{hyperref}
\let\labelindent\relax
\usepackage{enumitem}
\usepackage{xcolor}
\usepackage{xspace}
\usepackage{tikz}
\usetikzlibrary{patterns}
\usepackage{tabularx}
\usepackage{makecell} %
\usepackage{array} %
\usepackage[export]{adjustbox} %
\usepackage{subcaption} %

\usepackage{algorithm}
\usepackage{algorithmicx}
\usepackage{algpseudocode}
\usepackage{booktabs}
\usepackage{balance}
\usepackage{graphicx}
\usepackage{multirow}
\usepackage[font=small]{caption}
\usepackage{pgfplots}
\pgfplotsset{compat=1.18}
\usepackage{arydshln}
\usepackage{etoc}

\usepackage{CJKutf8}

\newcommand\methodname{GS-Playground}
\newcommand{\method}{\texttt{\methodname}\xspace}

\newcommand{\affilmark}[1]{\textsuperscript{#1}}
\newcommand{\equalmark}{\textsuperscript{*}}
\newcommand{\advisemark}{\textsuperscript{\ensuremath{\dagger}}}

\begin{document}

\setcounter{tocdepth}{-1}

\title{GS-Playground: A High-Throughput Photorealistic Simulator for Vision-Informed Robot Learning}

\author{\IEEEauthorblockN{
Yufei Jia\affilmark{1}\equalmark,
Heng Zhang\affilmark{2}\equalmark,
Ziheng Zhang\affilmark{3}\equalmark,
Junzhe Wu\affilmark{1}\equalmark,
Mingrui Yu\affilmark{1}\equalmark,\\
Zifan Wang\affilmark{5},
Dixuan Jiang\affilmark{6},
Zheng Li\affilmark{5},
Chenyu Cao\affilmark{7},
Zhuoyuan Yu\affilmark{3},
Xun Yang\affilmark{5},
Haizhou Ge\affilmark{1},\\
Yuchi Zhang\affilmark{4},
Jiayuan Zhang\affilmark{4},
Zhenbiao Huang\affilmark{8},
Tianle Liu\affilmark{3},
Shenyu Chen\affilmark{9},
Jiacheng Wang\affilmark{3},
Bin Xie\affilmark{3},\\
Xuran Yao\affilmark{4},
Xiwa Deng\affilmark{2},
Guangyu Wang\affilmark{1},
Jinzhi Zhang\affilmark{1},
Lei Hao\affilmark{10},
Zhixing Chen\affilmark{1},
Yuxiang Chen\affilmark{11},\\
Anqi Wang\affilmark{4},
Hongyun Tian\affilmark{3},
Yiyi Yan\affilmark{4},
Zhanxiang Cao\affilmark{12},
Yizhou Jiang\affilmark{1},
Hanyang Shao\affilmark{4},
Yue Li\affilmark{4},
Lu Shi\affilmark{1},\\
Wei Sui\affilmark{13},
Hanqing Cui\affilmark{2},
Yusen Qin\affilmark{13},
Ruqi Huang\affilmark{1}\advisemark,
Bokui Chen\affilmark{1}\advisemark,
Lei Han\affilmark{4}\advisemark,
Tiancai Wang\affilmark{3}\advisemark,
Guyue Zhou\affilmark{1}\advisemark\\[0.6em]}

\IEEEauthorblockA{
\affilmark{1}THU,
\affilmark{2}Motphys,
\affilmark{3}Dexmal,
\affilmark{4}DISCOVER Robotics,
\affilmark{5}HKUST(GZ),
\affilmark{6}BIT,\\
\affilmark{7}Galbot,
\affilmark{8}NUS,
\affilmark{9}HITSZ,
\affilmark{10}XJTU,
\affilmark{11}NJU,
\affilmark{12}SJTU,
\affilmark{13}D-Robotics}
\IEEEauthorblockA{
\equalmark Equal contribution.
\advisemark Advising.
Correspondence to: Yufei Jia \texttt{<jyf23@mails.tsinghua.edu.cn>}.}}

\let\oldtwocolumn\twocolumn
\renewcommand\twocolumn[1][]{%
    \oldtwocolumn[{#1}{
    \vspace{-5pt}
    \begin{center}
        \includegraphics[width=1.0\textwidth]{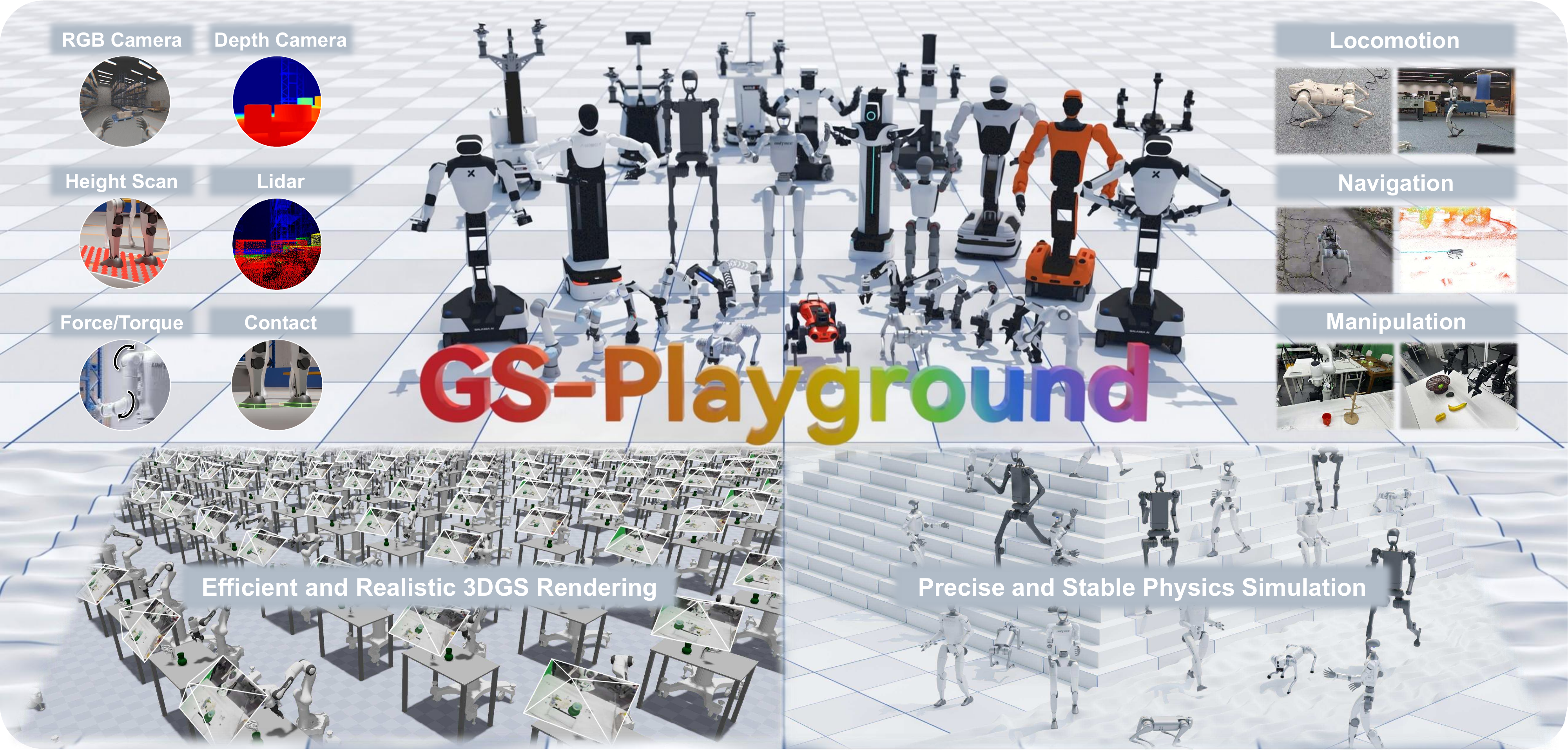}
        \captionof{figure}{\textbf{\method Overview.} It integrates photorealistic 3D Gaussian Splatting with high-performance parallel physics, achieving over $10^4$ aggregate FPS across 2048 environments at $640 \times 480$ resolution on a single GPU. We provide comprehensive sensor suites (Contact, Vision, LiDAR) and support a wide range of robotic embodiments and learning tasks, including locomotion, navigation, and manipulation.}
        \label{fig:teaser}
    \end{center}
    \vspace{3pt}
    }]
}

\maketitle

\begin{abstract}

Embodied AI research is undergoing a shift toward vision-centric perceptual paradigms. While massively parallel simulators have catalyzed breakthroughs in proprioception-based locomotion, their potential remains largely untapped for vision-informed tasks due to the prohibitive computational overhead of large-scale photorealistic rendering. Furthermore, the creation of simulation-ready 3D assets heavily relies on labor-intensive manual modeling, while the significant sim-to-real physical gap hinders the transfer of contact-rich manipulation policies. To address these bottlenecks, we propose \method, a multi-modal simulation framework for high-throughput, photorealistic robot learning in interactive rigid-body settings. We develop MotrixSim, a novel high-performance parallel physics engine, specifically designed to integrate with a batch 3D Gaussian Splatting (3DGS) rendering pipeline to ensure high-fidelity synchronization. Our system achieves an aggregate rendering throughput of $\mathbf{10^4}$ \textbf{FPS} across 2048 environments at $\mathbf{640 \times 480}$ resolution, significantly lowering the barrier for large-scale visual RL. Additionally, we introduce a \textbf{Real2Sim} workflow that reconstructs photorealistic and memory-efficient simulation-ready environments, reducing the manual effort required to create complex scenes for robot interaction. Extensive experiments on locomotion, navigation, and manipulation demonstrate that \method enables high-throughput photorealistic RL and robust Sim2Real transfer for interactive rigid-body robot learning. 
Project homepage: \url{https://gsplayground.github.io}.

\end{abstract}

\section{Introduction}
\label{sec:introduction}

\begin{table*}[t]
    \centering
    \caption{\textbf{Comparison of physical and perceptual capabilities across parallel robotics simulators.}}
    \label{tab:simulators}
    \resizebox{\textwidth}{!}{
    \renewcommand{\arraystretch}{1.1} 
    \setlength{\aboverulesep}{0pt}
    \setlength{\belowrulesep}{0pt}
    \begin{tabular}{c|cccc|cc|ccc|cc}
        \toprule
        \multirow{3}{*}{\textbf{Simulators}} & 
        \multirow{3}{*}{\makecell{\textbf{Physics}\\\textbf{Engine}}} & 
        \multirow{3}{*}{\makecell{\textbf{Batch}\\\textbf{Physics}}} & 
        \multirow{3}{*}{\makecell{\textbf{VRAM}\\\textbf{Usage}}} & 
        \multirow{3}{*}{\makecell{\textbf{Integrated}\\\textbf{Batch IK}}} & 
        \multirow{3}{*}{\makecell{\textbf{Batch}\\\textbf{Renderer}}} & 
        \multirow{3}{*}{\makecell{\textbf{Batch}\\\textbf{Render}\\\textbf{Fidelity}}} & 
        \multirow{3}{*}{\makecell{\textbf{3DGS}\\\textbf{Env.}\\\textbf{Num.}}} & 
        \multirow{3}{*}{\makecell{\textbf{Dynamic}\\\textbf{3DGS }\\\textbf{Scene}}} & 
        \multirow{3}{*}{\makecell{\textbf{3DGS}\\\textbf{Render} \\\textbf{FPS}}} & 
        \multirow{3}{*}{\makecell{\textbf{Startup}\\\textbf{Speed}}} & 
        \multirow{3}{*}{\makecell{\textbf{Physics}\\\textbf{Cross}\\\textbf{Platform}}} \\ 
        & & & & & & & & & & & \\ %
        & & & & & & & & & & & \\ %
        \hline
        MuJoCo/MJX~\cite{todorov2012mujoco}    & Brax/MJX     & CPU/GPU & {$\star\star$}                & {$\times$}     & Madrona    & +   & -          & -              & - & +    & L     \\
        IsaacLab~\cite{mittal2025isaac}        & PhysX5       & GPU     & {$\star\star\star\star\star$} & {$\checkmark$} & omni.RTX   & ++  & -          & -              & - & ++   & L     \\
        ManiSkill~\cite{taomaniskill3}         & PhysX5       & GPU     & {$\star\star\star$}           & {$\checkmark$} & Vulkan SBR & +   & -          & -              & - & +++  & W/L   \\
        Genesis~\cite{Genesis}                 & Taichi       & GPU     & {$\star\star$}                & {$\checkmark$} & Madrona    & +   & -          & -              & - & ++   & W/L/M \\
        \hline
        RoboGSim~\cite{li2024robogsim}         & PhysX4       & -       & -                             & {$\times$}     & -          & +++ & 1          & {$\checkmark$} & $\sim$ 100 & ++   & L     \\
        DISCOVERSE~\cite{jia2025discoverse}    & MuJoCo       & -       & -                             & {$\times$}     & -          & +++ & $1 \sim 4$ & {$\checkmark$} & $\sim$ 650 & ++   & L     \\
        GSWorld~\cite{jiang2025gsworld}        & PhysX5       & -       & -                             & {$\times$}     & -          & +++ & 1          & {$\checkmark$} & -          & ++   & L     \\
        GaussGym~\cite{escontrela2025gaussgym} & PhysX4       & GPU     & {$\star$}                     & {$\times$}     & GSplat     & +++ & Up To 4096 & {$\times$}     & -          & ++   & L     \\
        \hline
        \textbf{\method} & MotrixSim     & CPU/GPU & -                             & {$\checkmark$} & BatchSplat & +++ & Up To 4096 & {$\checkmark$} & $\sim\textbf{10k}$ & ++++ & W/L/M \\
        \bottomrule
    \end{tabular}
    }
    \vspace{-1mm}
    \footnotesize
    \begin{flushleft}
    \textbf{Note:} 
    1. \textbf{Batch Physics}: Indicates supported hardware for batched simulation. 
    2. \textbf{VRAM Usage}: The number of `$\star$' indicates higher VRAM consumption. In headless mode, only physics simulation overhead is considered. 
    3. \textbf{Batch Render Fidelity}: The number of `+' represents higher visual fidelity. 
    4. \textbf{3DGS Render FPS}: Aggregate throughput across 2048 environments at 640$\times$480 resolution with an NVIDIA RTX 4090 GPU and an Intel i9-14900K CPU.
    5. \textbf{Startup Speed}: More `+' means faster startup times.
    6. \textbf{Physics Cross Platform}: W: Windows, L: Linux, M: macOS.
    \end{flushleft}
    \vspace{-7mm}
\end{table*}

Vision serves as the most information-rich modality for robotic perception of the environment. Recently, significant progress has been made in learning policies for quasi-static manipulation and navigation directly from real-world visual data \cite{kim2024openvla,zhou2025vision,black2024pi_0,zhai2025vision,li2025simplevla,xue2025opening,he2025viral}. However, tasks involving complex dynamics and contacts—such as locomotion and contact-rich manipulation—require large-scale parallel simulation for effective reinforcement learning (RL). 
While current massively parallel simulators have revolutionized proprioception-based learning \cite{makoviychuk2021isaac, mittal2025isaac, zakka2025mujoco, taomaniskill3, Genesis}, they often struggle to reconcile visual fidelity with rendering efficiency, limiting the application of large-scale, vision-informed policy learning. 
We attribute this gap to two primary limitations:
\begin{itemize}
    \item \textbf{Prohibitive Rendering Overhead}: Existing simulation pipelines face severe scalability bottlenecks when integrating high-resolution rendering. The exorbitant computational cost of photorealistic rendering creates intense resource contention with policy learning, frequently resulting in Out-of-Memory (OOM) failures. Consequently, these hardware constraints force a compromise between visual fidelity and simulation throughput, restricting the scale and efficacy of vision-based training. Recent 3DGS-based robotic systems such as GSWorld~\cite{jiang2025gsworld} and RoboGSim~\cite{li2024robogsim} attach photorealistic rendering to manipulation pipelines, but run as single-environment closed-loop evaluators rather than batch-rendered, on-policy training engines.

    \item \textbf{Laborious Asset Synthesis}: Constructing simulation assets that combine visual fidelity with robot-interaction compatibility remains a persistent challenge. While 3D reconstruction has advanced significantly, converting these representations into ``sim-ready" assets compatible with high-frequency rigid-body physics and memory-efficient rendering remains difficult and laborious. Vid2Sim~\cite{xie2025vid2sim} and Video2Game~\cite{xia2024video2game} reconstruct navigable scenes or playable games from a single video, but target agent navigation and graphics rather than robot-object interaction. This highlights the critical need for a pipeline capable of rapidly transforming individual real-world scenes into high-fidelity simulation environments for robot interaction.
\end{itemize}

To bridge the gap, we introduce \method, a robot-learning simulation framework that harmonizes high-throughput rigid-body physics simulation and high-fidelity visual rendering (Figure \ref{fig:teaser}). Our platform maintains the precision and stability required for physics simulation while providing the rendering efficiency necessary for large-scale, vision-informed policy training and sim-to-real transfer. 
Our core contributions are as follows:

\begin{enumerate}
    \item \textbf{Embodied Robot Simulation Platform:} We develop MotrixSim, a ground-up, cross-platform (Windows, Linux, and macOS) parallel physics engine that supports both GPU and CPU backends. This architecture provides high-fidelity rigid-body dynamics and comprehensive sensor integration (including RGB cameras, LiDAR, and force/contact sensors) across diverse robot embodiments (such as quadrupeds, humanoids, and manipulators).
    This platform facilitates a flexible development workflow from local prototyping to massively parallel training.
    
    \item \textbf{Memory-Efficient Batch 3DGS Rendering:} To mitigate the memory bottlenecks in large-scale photorealistic simulation, we introduce a specialized point-pruning strategy optimized for rigid-body environments. This approach enables a memory-efficient rendering architecture capable of a breakthrough aggregate throughput of $\mathbf{10^4}$ \textbf{FPS} across 2048 environments at $\mathbf{640 \times 480}$ resolution on a single GPU, significantly expanding the scale of vision-based reinforcement learning.
    
    \item \textbf{``Sim-Ready" Real2Sim Workflow:} 
    We present a targeted pipeline that streamlines the conversion of individual real-world scenes into simulation-ready rigid-body environments with photorealistic rendering and task-configurable physical parameters. Our workflow significantly reduces the laborious manual effort usually required to create complex "sim-ready" assets, enabling the rapid population of diverse simulation environments.
\end{enumerate}

We validate \method through a rigorous experimental regime. First, we demonstrate that our physics engine matches state-of-the-art simulators in accuracy and efficiency, with enhanced stability in specific contact-rich scenarios. Next, we confirm our rendering pipeline's ability to provide high-fidelity feedback at unprecedented speeds. Finally, we benchmark the framework on a diverse suite of tasks---including quadrupedal locomotion, humanoid control, navigation, and robotic manipulation---across both state-based and vision-driven reinforcement learning.
We summarize the key features of \method alongside other state-of-the-art simulators in Table \ref{tab:simulators}.
We will release the framework and synthesized \textit{Bridge-GS} dataset to empower the research community.

\section{Related Work}
\label{sec:related_work}

\subsection{Massive Parallelism in Simulation}
Massive parallel physical simulation has emerged as the indispensable infrastructure for efficient robot learning, particularly for locomotion and contact-rich manipulation \cite{hwangbo2019learning,kumar2021rma,siekmann2021blind,margolis2023walk,choi2023learning,zhuang2023robot,margolis2024rapid,wang2024arm,he2025asap,wang2025omni, zhang2025keep, cao2025learning}. 
Milestone platforms such as Isaac Gym \cite{makoviychuk2021isaac}, Isaac Lab \cite{mittal2025isaac}, and Genesis \cite{Genesis} have revolutionized the throughput of sample collection by instantiating tens of thousands of parallel environments on GPUs.
However, existing frameworks typically prioritize physical throughput over perceptual fidelity.
Rendering architectures often diverge into two extremes: they either favor high sampling rates via streamlined rasterization (e.g., Madrona \cite{shacklett2023extensible}, ManiSkill3 \cite{taomaniskill3}) or prioritize cinematic photorealism via computationally expensive ray-tracing (e.g., Isaac Lab \cite{mittal2025isaac}).
The simultaneous achievement of massive parallel throughput with high-fidelity, photorealistic rendering remains a critical bottleneck for scaling vision-centric robot learning.

\begin{figure*}[t]
    \centering
    \includegraphics[width=\linewidth]{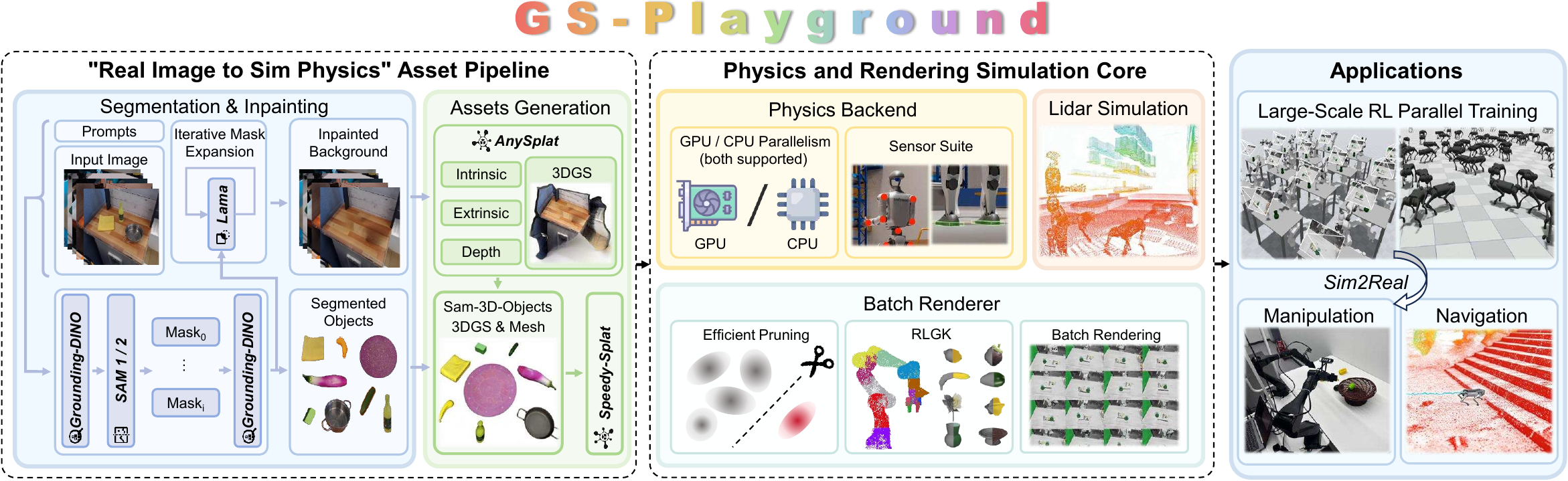}
    \vspace{-5mm}
    \caption{\textbf{GS-Playground System Architecture.} \textbf{Left:} an automated Real2Sim pipeline that constructs simulation-ready visual and geometric assets from RGB inputs via object segmentation, background inpainting, and 3DGS/mesh reconstruction. \textbf{Middle:} a physics and rendering simulation core with CPU/GPU physics backends, integrated sensor and LiDAR simulation, and batch-optimized 3DGS rendering with pruning and rigid-link kinematics. \textbf{Right:} downstream applications including manipulation, navigation, and large-scale parallel reinforcement learning.}
    \label{fig:system}
    \vspace{-5mm}
\end{figure*}

\subsection{Vision-Centric Robot Learning}

Vision serves as the most information-rich modality for robotic perception. Recently, significant progress has been made in learning policies for quasi-static manipulation and navigation directly from real-world visual data, exemplified by Vision-Language-Action (VLA) models \cite{kim2024openvla, zhou2025vision, black2024pi_0} and Vision-Language-Navigation (VLN) models \cite{cheng2024navila, cai2025navdp, zhang2024uninavid, zhang2025embodied}. However, tasks involving complex dynamics and intermittent contacts rely heavily on simulation-based reinforcement learning (RL) to acquire skills in an unsupervised manner.

Early attempts to incorporate visual inputs into RL were often constrained by conventional, small-scale simulations \cite{akkaya2019solving,matas2018sim,zhu2020robosuite}, where low simulation throughput hindered the stable acquisition of complex skills.
While recent advancements in massive parallel simulation have enabled sophisticated policy optimization for locomotion and dexterous manipulation, these frameworks primarily rely on proprioceptive states or point clouds due to the prohibitive computational overhead and limited fidelity of visual rendering \cite{chen2023visual,li2025clone,jiang2025robust,yin2025dexteritygen}.
Furthermore, to mitigate the sim-to-real gap, recent attempts on vision-based RL methods often necessitate extensive visual randomization of textures, lighting, and backgrounds \cite{he2025viral,xue2025opening}. This, however, imposes a substantial burden on GPU resources, significantly raising the computational threshold for research in vision-informed robot learning.

\subsection{Gaussian Splatting and Real-to-Sim Reconstruction}
Building simulators highly consistent with the real world relies on high-quality visual rendering and precise physical modeling. Recently, Gaussian Splatting (3DGS) \cite{kerbl20233d} has rapidly developed as an emerging scene reconstruction method, enabling photorealistic real-time rendering from arbitrary viewpoints. Its unique representation strikes a balance between visual fidelity and memory efficiency, making it particularly suitable for resource-constrained simulation environments. Recent research has significantly improved the capability to reconstruct renderable models from real scenes in terms of reconstruction paradigms \cite{wang2025vggt,mescheder2025sharp,jiang2025anysplat}, model efficiency \cite{hanson2025pup,fang2024mini,hanson2025speedy}, and generative Gaussian models \cite{chen2025sam,lin2025diffsplat}. 

However, these methods generally struggle to directly meet the requirements of robotic simulators for sim-ready scenes. Existing works indicate that 3DGS-based rendering holds significant potential in robotics, enhancing sim-to-real transfer for vision-based policies \cite{li2024robogsim,han2025re,qureshi2025splatsim}, augmenting training datasets \cite{barcellona2024dream,lou2025robo,yu2025real2render2real,yang2025novel}, and supporting real-to-sim evaluation pipelines \cite{jiang2025phystwin,abou2025real,jiang2025gsworld,zhang2025real}. While GaussGym \cite{escontrela2025gaussgym} pioneered the application of 3DGS in RL, our work extends this capability to contact-rich manipulation and larger-scale parallel throughput, providing a more versatile foundation for diverse vision-informed robot learning tasks.

\section{System Design}
\label{sec:method}

\subsection{System Overview}
\label{sec:overview}

Figure~\ref{fig:system} illustrates the architecture of \method, which comprises three core tiers: (1) MotrixSim, a \textbf{high-performance parallel physics engine} supporting both GPU and CPU backends; (2) a \textbf{memory-efficient batch 3DGS renderer} optimized via point-pruning; and (3) an \textbf{automated Real2Sim pipeline} for rapid scene synthesis. These modules are seamlessly integrated with a multi-modal sensor suite to provide the high-throughput, photorealistic feedback necessary for vision-centric robot learning.

\textbf{Data Flow.} 
The simulation loop begins with the physics engine, which advances the world state using a velocity-impulse formulation. The updated rigid-body poses are synchronized with the Batch 3DGS Renderer through \textit{Rigid-Link Gaussian Kinematics} (RLGK), enabling zero-overhead updates of visual clusters. The renderer produces photorealistic RGB images and depth maps, while the sensor suite provides LiDAR point clouds and high-dimensional contact data (including multi-point forces and torques). These modalities form the observation vector for end-to-end policy learning.

\textbf{Asset Creation.} 
Orthogonal to the loop, the \textit{Real2Sim Pipeline} streamlines the conversion of individual real-world captures into simulation-ready assets. By performing automated segmentation, reconstruction, and sub-millimeter pose alignment, this pipeline produces photorealistic rigid-body assets for robot interaction.

\textbf{Development Workflow.} 
A key design principle is the flexibility of the development-to-training workflow. Our core physics engine is cross-platform (Windows, Linux, macOS), facilitating rapid local debugging and prototyping on various workstations. For large-scale training, the system leverages an optimized CUDA-based rendering pipeline on Linux, achieving a breakthrough aggregate throughput of $10^{4}$ FPS across 2048 environments at $640\times480$ resolution by utilizing specialized point-pruning to balance visual fidelity and memory consumption.

\subsection{Physics Solver Formulation}
\label{sec:physics}
In robotic manipulation, the choice of constraint formulation directly impacts simulation fidelity. Optimization-centric solvers that rely on regularized soft contacts tend to produce visually smooth but physically 'spongy' interactions, where heavy payloads may exhibit gradual drift due to residual forces.
MotrixSim utilizes a velocity-impulse formulation in generalized coordinates and implements strict complementarity with explicit velocity clamping at the friction limits. This approach sacrifices the smoothness of the gradients in exchange for geometric precision, allowing for the simulation of rigid bodies that can maintain perfect static equilibrium and allows for high constraint stiffness and large simulation time steps. This makes the engine particularly suitable for engineering applications where stability and exact constraint satisfaction are paramount.

The discretized dynamics equation for generalized coordinates $\mathbf{q} \in \mathbb{R}^n$ and velocities $\mathbf{v} \in \mathbb{R}^n$ over a time step $h$ is formulated as:
\begin{equation}
    \mathbf{M}(\mathbf{v}^+ - \mathbf{v}) = \mathbf{J}_e^T \boldsymbol{\lambda}_e^+ + \mathbf{J}_n^T \boldsymbol{\lambda}_n^+ + h(\boldsymbol{\tau}_{ext} - \mathbf{c})
\end{equation}
where $\mathbf{M}$ is the mass matrix, $\mathbf{J}_e$ and $\mathbf{J}_n$ are the Jacobians for equality and inequality constraints, respectively, and $\boldsymbol{\lambda}$ denotes the constraint impulses. The term $\mathbf{c}$ accounts for Coriolis and centrifugal forces.
We incorporate soft constraints by defining an implicit impulse relation $\boldsymbol{\lambda}^+ = f(\mathbf{u}^+; \mathbf{x}, h)$ and linearizing it via a first-order Taylor expansion at the current velocity $\mathbf{u}$:
\begin{equation}
    \boldsymbol{\lambda}^+ \approx f(\mathbf{u}) + \frac{\partial f}{\partial \mathbf{u}}(\mathbf{u}^+ - \mathbf{u})
\end{equation}
By defining the positive definite compliance matrix $\mathbf{C} = (-\frac{\partial f}{\partial \mathbf{u}})^{-1}$ and the bias term $\boldsymbol{\zeta} = \mathbf{u} + \mathbf{C}f(\mathbf{u})$, we obtain the standardized compliance form:
\begin{equation}
    \mathbf{u}^+ = -\mathbf{C}\boldsymbol{\lambda}^+ + \boldsymbol{\zeta}
\end{equation}
By substituting this velocity relation into the constraint space and eliminating the equality constraints $\boldsymbol{\lambda}_e$ via the Schur complement method, we obtain a reduced linear system for the inequality constraints $\boldsymbol{\lambda}_n$:
\begin{equation}
    \mathbf{u}_n^+ = \mathbf{A} \boldsymbol{\lambda}_n^+ + \mathbf{b}
\end{equation}
The system matrix $\mathbf{A}$ and the right-hand side vector $\mathbf{b}$ are explicitly given by:
\begin{align}
    \mathbf{A} &= \mathbf{J}_n \mathbf{M}^{-1} \mathbf{J}_n^T - \mathbf{J}_n \mathbf{M}^{-1} \mathbf{J}_e^T (\mathbf{W}_{ee} + \mathbf{C}_e)^{-1} \mathbf{J}_e \mathbf{M}^{-1} \mathbf{J}_n^T \\
    \mathbf{b} &= \mathbf{J}_n \tilde{\mathbf{v}} + \mathbf{J}_n \mathbf{M}^{-1} \mathbf{J}_e^T (\mathbf{W}_{ee} + \mathbf{C}_e)^{-1} (\boldsymbol{\zeta}_e - \mathbf{J}_e \tilde{\mathbf{v}})
\end{align}
where $\mathbf{W}_{ee} = \mathbf{J}_e \mathbf{M}^{-1} \mathbf{J}_e^T$ is the effective inverse mass matrix of the equality constraints. The term $(\mathbf{W}_{ee} + \mathbf{C}_e)$ is guaranteed to be invertible as it is the sum of a positive semi-definite matrix and a positive definite matrix.

The solver resolves contact and friction as a Mixed Complementarity Problem (MCP). The impulse vector $\boldsymbol{\lambda}_n$ comprises normal components $\boldsymbol{\lambda}_{\perp}$ and frictional components $\boldsymbol{\lambda}_{\parallel}$. The solution must satisfy the bounds defined by the Coulomb friction model:
\begin{equation}
    \begin{cases}
    w_i \ge 0, \quad \text{if } \lambda_{i}^+ = l_i \\
    w_i = 0, \quad \;\; \text{if } l_i < \lambda_{i}^+ < u_i \\
    w_i \le 0, \quad \text{if } \lambda_{i}^+ = u_i
    \end{cases}
\end{equation}
where $w_i = [(\mathbf{A} + \mathbf{C}_n)\boldsymbol{\lambda}_n^+ + (\mathbf{b} - \boldsymbol{\zeta}_n)]_i$. For normal contact, the bounds are $[0, \infty)$; for friction, the bounds are $[-\mu \lambda_{\perp}^+, \mu \lambda_{\perp}^+]$. This formulation is solved efficiently using a Projected Gauss-Seidel (PGS) solver, ensuring stable friction behavior while accommodating both rigid and compliant contact interactions.

This framework demonstrates high extensibility. It supports various physical constraints, including MJCF-defined contact models (e.g., parameters \texttt{solref}, \texttt{solimp}), tendons, and actuators. Integrating a new constraint type simply requires defining its impulse-state relationship $\boldsymbol{\lambda}(\mathbf{x}, \mathbf{u})$ and the corresponding Jacobian $\mathbf{J}$.
Additionally, to achieve real-time performance in large-scale scenarios with massive contacts, we implemented two key engineering optimizations:

1) \textbf{Parallelization via Constraint Islands:} Leveraging the spatial locality of physical interactions, we dynamically construct a constraint dependency graph at each time step. By analyzing the connectivity of the graph, the rigid body system is partitioned into disjoint sets of interacting bodies, termed ``Constraint Islands.'' Since the Linear Complementarity Problems (LCPs) for these islands are mathematically independent, they are dispatched to multi-core CPU threads for parallel solving, ensuring linear performance scaling with scene complexity.

2) \textbf{Warm-Starting with Temporal Coherence:} We exploit the temporal coherence of physical processes by implementing a Contact Manifold Tracking system. This system persists contact constraints across simulation frames. Instead of initializing the Projected Gauss-Seidel (PGS) solver with zero vectors, we warm-start the solver using the converged impulses $\lambda_{t-1}$ from the previous frame as the initial guess $\lambda_{\text{initial}}$. This strategy significantly accelerates convergence, typically reducing the required PGS iterations from over 50 to fewer than 10 for stable stacking tasks.

\subsection{Batch Renderer Optimization}
\label{sec:renderer}

Rendering thousands of high-fidelity 3DGS scenes simultaneously presents a significant memory challenge. To optimize memory usage while maintaining visual fidelity, we propose several key advancements as follows:

\textbf{Efficient Pruning Strategy}. We adopt an efficient pruning strategy inspired by recent works \cite{hanson2025pup,fang2024mini,hanson2025speedy} to reduce the number of Gaussians while preserving visual quality for robot-learning workloads. For static scenes, our pruned representation keeps 30\% of the original Gaussians while maintaining high PSNR and SSIM (Table~\ref{tab:visual_loss_com}); for dynamic objects and robot-linked assets, pruning can be more aggressive depending on visual complexity. This reduction significantly lowers VRAM usage and improves rendering speed, making the renderer well-suited for large-scale, high-fidelity simulations.

\textbf{Throughput Scaling}. We build our Batch‑3DGS rendering pipeline on top of gsplat \cite{ye2025gsplat} and introduce simulation-oriented system optimizations for large-scale parallel environments. With these optimizations, we can render up to 2048 scenes at a resolution of $640\times480$ with a total throughput of up to 10,000 FPS; in single-environment rendering of the same scene, our pruned 3DGS reaches 648 FPS versus 125 FPS for vanilla 3DGS. This scaling significantly improves throughput per unit compute, supporting large‑batch training workflows.

\textbf{Rigid-Link Gaussian Kinematics (RLGK)}. To ensure temporal consistency and eliminate visual artifacts in dynamic scenarios, we introduce Rigid‑Link Gaussian Kinematics (RLGK), which binds clusters of 3D Gaussians to corresponding rigid bodies in the physics engine. This coupling ensures that visual representations move synchronously with their physical counterparts, enabling “zero‑overhead” updates and artifact‑free dynamic rendering during fast motion or contact events.

\subsection{Real2Sim Asset Pipeline}
\label{sec:real2sim}

Users can create a set of simulation-ready assets from a single RGB image via our Real2Sim pipeline.
The pipeline reconstructs 3D representations (3DGS/mesh), estimates geometric pose and scale, and supports task-level physical parameters for collision modeling and manipulation learning.
In our experiments, mass, friction, and inertia are assigned through task configuration, manual tuning, or domain randomization rather than estimated from the input image; the modular pipeline can incorporate system identification when interaction data is available.

\textbf{Objects Segmentation and Background Inpainting}.
We present an automated pipeline for object segmentation and background inpainting from a single RGB image. Objects are detected using Grounding DINO \cite{liu2023grounding} and segmented with SAM1/SAM2 \cite{kirillov2023segany,ravi2024sam2} under a prompt-wise independent detection scheme, enabling explicit tracking of instance–label associations. To mitigate unreliable semantic similarity in open-vocabulary detection, we de-duplicate instances using visual criteria only: mask IoU for general redundancy and a dual-criterion rule based on mask inclusion and boundary overlap to correct over-segmentation. Instance selection is prioritized by a composite confidence score combining detection and segmentation confidence.
Occluded object regions are recovered through an iterative mask expansion process coupled with sequential inpainting. Objects are removed one at a time, and after each inpainting step, the scene is re-detected to identify newly exposed regions that are spatially adjacent and label-consistent with existing instances under a bounded area growth constraint. Background inpainting is performed sequentially using LaMa \cite{suvorov2021resolution} to ensure stable reconstruction.

\textbf{Assets Generation}.
For object-level assets, we apply SAM-3D \cite{chen2025sam} to the original RGB image together with the object mask ($M_{obj}$) to reconstruct its 3DGS and mesh, and to estimate its pose and scale.
For scene-level assets, the inpainted background is processed by AnySplat \cite{jiang2025anysplat} to generate the background 3DGS, depth map ($D_{bg}$), as well as the camera intrinsics and extrinsics.
To align object- and scene-level assets, we first transform the object such that its rendered depth map ($D_{obj}$) aligns with the background depth ($D_{bg}$). The object is then scaled so that the area of its rendered mask matches the original object mask ($M_{obj}$), measured by pixel occupancy.
To reduce memory footprint for downstream robotic tasks, we apply Speedy- splat \cite{hanson2025speedy} for 3DGS pruning.

\begin{table}[t]
    \centering
    \caption{\small \textbf{Comparison of LiDAR simulation capabilities}.}
    \label{tab:lidar}
    \scalebox{1}{
        \begin{tabular}{lccc}
            \toprule
            \textbf{LiDAR Sensor Feature} & \textbf{Ours} & \textbf{IsaacSim} & \textbf{Gazebo} \\
            \midrule
            Rotating LiDAR Support    & \checkmark & \checkmark & \checkmark \\
            Solid-State LiDAR Support & \checkmark & \checkmark & \checkmark \\
            Non-repetitive scan LiDAR & \checkmark & {$\times$} & \checkmark \\
            Static Irregular Objects  & \checkmark & \checkmark & \checkmark \\
            Dynamic Irregular Objects & \checkmark & {$\times$} & {$\times$} \\
            Self-Occlusion            & \checkmark & {$\times$} & {$\times$} \\
            3DGS Representation       & \checkmark & {$\times$} & {$\times$} \\
            Massively Parallel        & \checkmark & \checkmark & {$\times$} \\
            \bottomrule
        \end{tabular}
    }
    \vspace{-5mm}
\end{table}

\subsection{User Interface and Ecosystem Design}
\method provides a rich set of features including multi-modal sensing, seamless ecosystem compatibility, and cross-platform support to facilitate robot development.

\textbf{Multi-modal Sensor Suite} 
Beyond standard RGB and depth streams, we integrate a high-performance Batch-LiDAR module utilizing ray-casting to generate high-fidelity point clouds and heightmap scanning (Table \ref{tab:lidar}). 
Additionally, our platform provides detailed contact information equivalent to MuJoCo, including multi-point contact forces, torques, and decomposed normal/tangential components.

\textbf{Ecosystem Compatibility and Cross-Platform Workflow}.
Prioritizing "zero-friction" migration, our API is compatible with the MuJoCo MJCF format, enabling rapid project migration. The physics engine features a cross-platform architecture (Windows/Linux/macOS), allowing local prototyping and debugging before deploying large-scale parallel training on Linux GPU clusters using Batch-3DGS and CUDA-optimized perception modules.

\section{Results}
\label{sec:results}

We evaluate \method through a comprehensive benchmark suite spanning visual and geometric fidelity, physics stability, manipulation proficiency, and locomotion capability. Our results demonstrate the platform's superiority in photorealistic rendering, massive parallel physics stepping, and effective Sim2Real transfer for both vision-based and contact-rich tasks.

\subsection{Physics Stability and Solver Robustness}

\begin{table}[t]
    \caption{\small \textbf{Qualitative comparison of contact dynamics}. The Newton’s Cradle case evaluates momentum transfer, while the Boston Dynamics Spot case validates base stability with a 10 ms timestep. }
    \label{tab:newton_cradle_and_boston_spot}
    \centering
    \begin{tabular}{>{\centering\arraybackslash}m{0.22\linewidth}ccc} 
        \toprule
        \textbf{Scenario} & \textbf{MuJoCo} & \textbf{IsaacSim} & \textbf{Ours} \\
        \midrule
        \shortstack{Newton's Cradle\\(pre-impact)} &
        \includegraphics[height=1.2cm, valign=m]{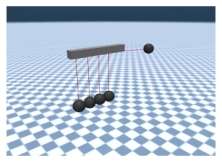} &
        \includegraphics[height=1.2cm, valign=m]{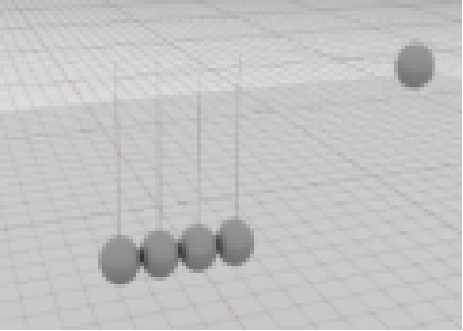} &
        \includegraphics[height=1.2cm, valign=m]{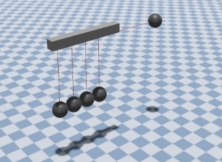} \\
        
        \shortstack{Newton's Cradle\\(post-impact)} &
        \includegraphics[height=1.2cm, valign=m]{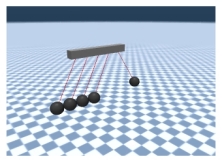} &
        \includegraphics[height=1.2cm, valign=m]{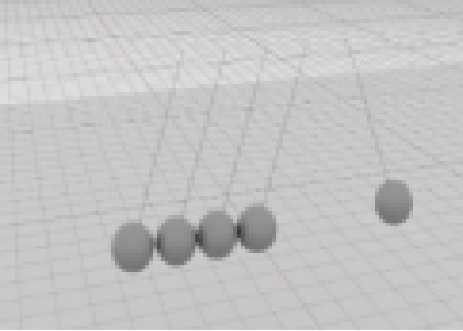} &
        \includegraphics[height=1.2cm, valign=m]{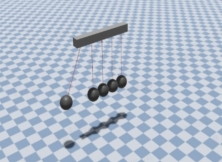} \\
        
        \midrule %
        
        \shortstack{Boston Spot\\(t=0s)} &
        \includegraphics[height=1.2cm, valign=m]{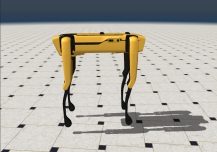} &
        \includegraphics[height=1.2cm, valign=m]{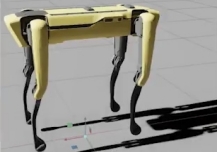} &
        \includegraphics[height=1.2cm, valign=m]{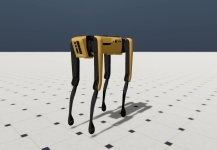} \\
        
        \shortstack{Boston Spot\\(t=2s)} &
        \includegraphics[height=1.2cm, valign=m]{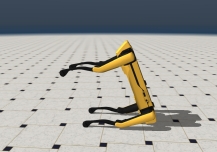} &
        \includegraphics[height=1.2cm, valign=m]{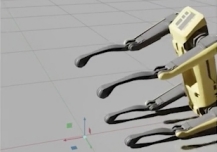} &
        \includegraphics[height=1.2cm, valign=m]{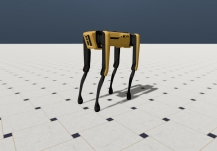} \\
        \bottomrule
    \end{tabular}
\end{table}

\textbf{Multi-Scenario Benchmarking.}
We conduct a multi-scenario stability study across multiple simulators to stress-test contact handling under challenging dynamics.

1) \textit{Hard Contact \& Momentum Conservation}: Using a Newton's Cradle setup with identical initial perturbations across engines, we evaluate long-horizon momentum transfer and dissipation under hard contacts (Table \ref{tab:newton_cradle_and_boston_spot} Top). \method preserves impact timing and swing amplitude with reduced energy bleed across repeated impacts, while MuJoCo exhibits stronger damping and phase drift.

2) \textit{Large Time-step Stability}: We evaluate base stability on a Boston Dynamics Spot model under physics-only stepping with a 10\,ms timestep, no control input, and identical initial pose (Table \ref{tab:newton_cradle_and_boston_spot} Bottom). \method exhibits smaller base displacement and reduced drift over time, suggesting more stable contact resolution under large time steps.

\begin{figure}[t]
    \centering
    \begin{subfigure}[c]{0.40\linewidth}
        \centering
        \includegraphics[width=\linewidth]{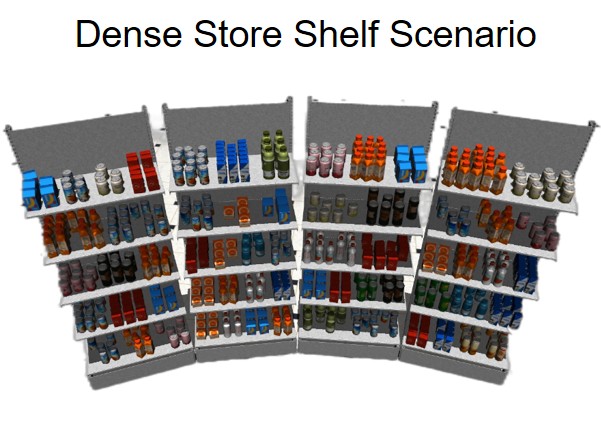}
        \label{fig:store_snapshot}
    \end{subfigure}
    \hfill %
    \begin{subfigure}[c]{0.57\linewidth}
        \centering
        \includegraphics[width=\linewidth]{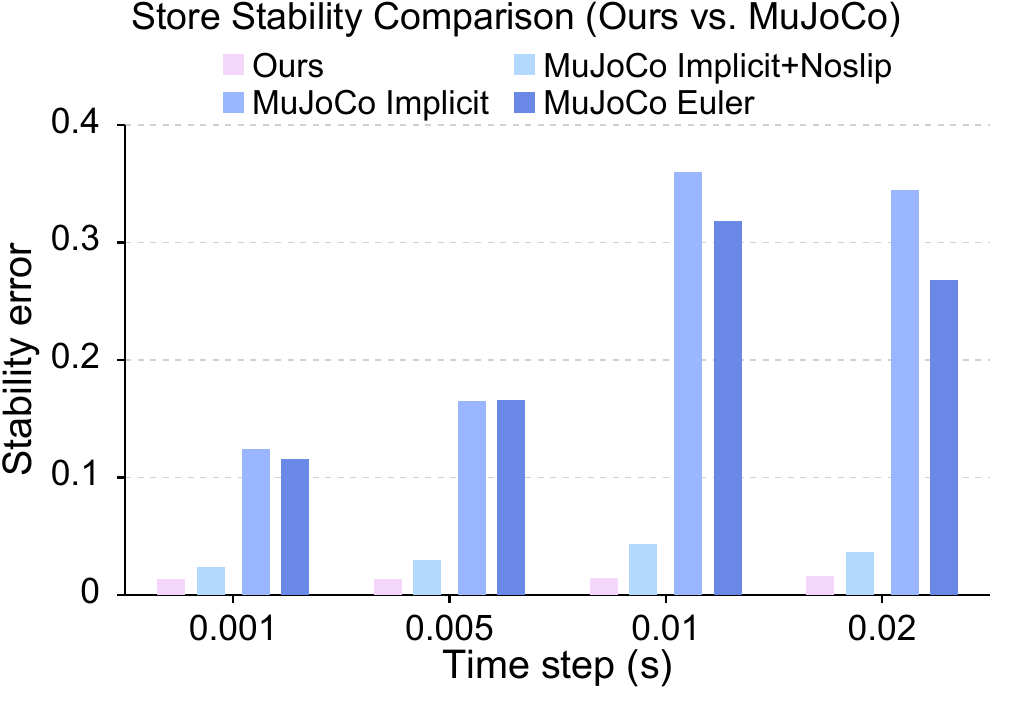}
        \label{fig:store_plot}
    \end{subfigure}
    \vspace{-7mm} %
    \caption{\textbf{Physics stability under complex multi-body interactions}. 
    (a) The dense store shelf scenario; (b) Stability error across time steps, computed over all objects in the scene with identical initial placements. The error is defined as $\sqrt{\Delta p^2 + \Delta \theta^2}$, where $\Delta p$ is mean positional drift (m) and $\Delta \theta$ is mean orientation drift (rad).}
    \vspace{-3mm}
    \label{fig:store}
\end{figure}

3) \textit{Complex Multi-Body Interactions}: In a dense store shelf scenario with stacked objects, we evaluate static stability under complex multi-contact constraints. While \method consistently converges to stable equilibria, MuJoCo exhibits characteristic jitter and contact-induced drift—common artifacts in high-density contact graphs (Fig. \ref{fig:store}).

\begin{figure}[t]
    \centering
    \includegraphics[width=\linewidth]{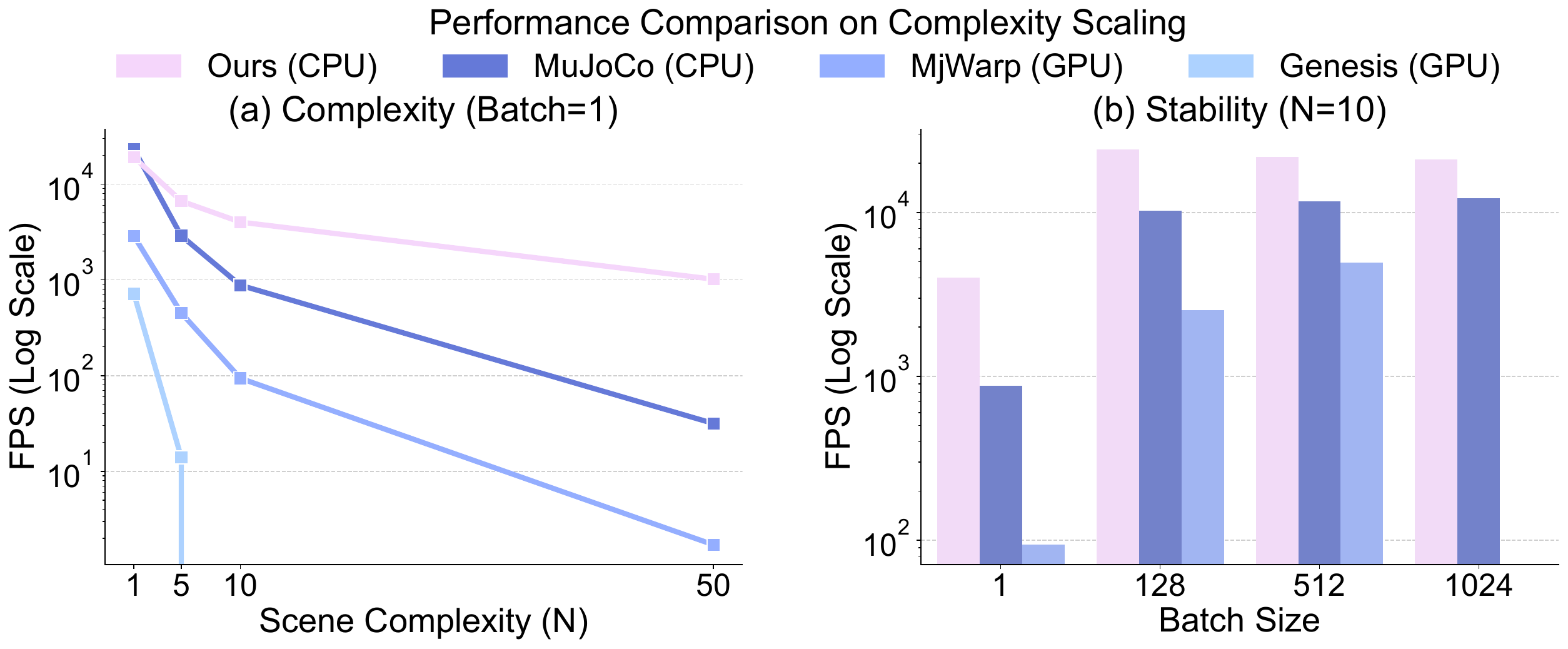}
    \caption{\textbf{Performance Comparison on Complexity Scaling.} 
    Left: As the complexity ($N$, the number of humanoid robots) in a single environment increases, the FPS advantage of our framework becomes increasingly pronounced.
    Right: At a complexity of $N=10$, our framework maintains FPS advantage with large batch sizes, where Genesis fails to reach convergence.
    }
    \label{fig:benchmark}
    \vspace{-5mm}
\end{figure}

\begin{figure*}[t!]
\centering
\includegraphics[width=\linewidth]{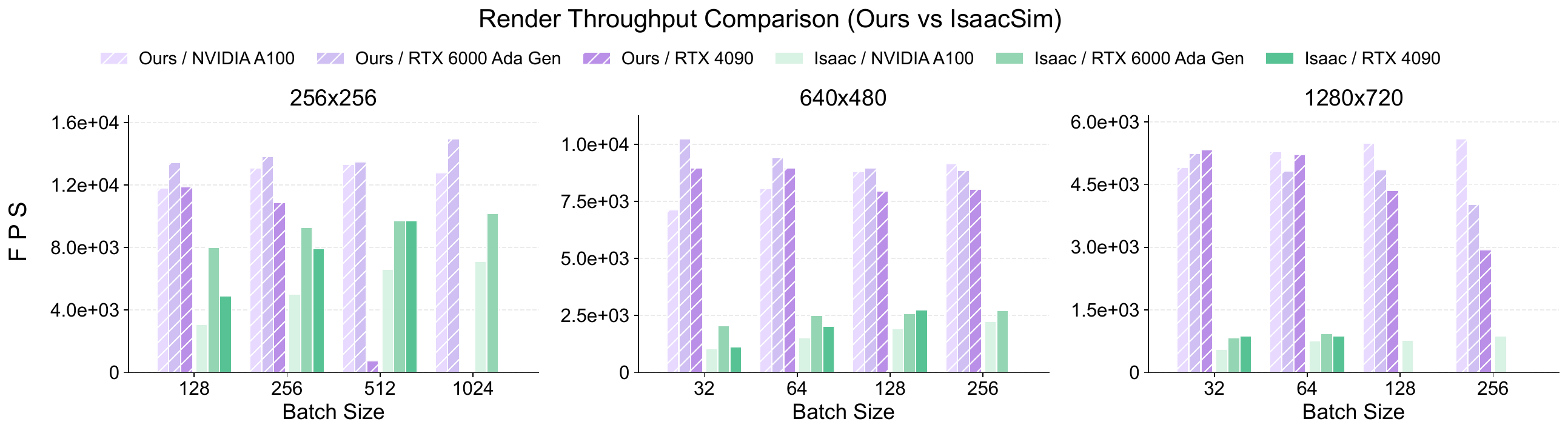}
\vspace{-0.6cm}
\caption{\textbf{Aggregate rendering throughput of GS-Playground vs. Isaac Sim's ray-tracing renderer across resolutions.} Isaac Sim relies on manual asset modeling and encounters Out-Of-Memory (OOM) exceptions at higher resolutions, while GS-Playground leverages automated asset generation from real-world captures. Evaluations span three GPU architectures: NVIDIA RTX 4090, RTX 6000 Ada Generation, and NVIDIA A100.}
\vspace{-4mm}
\label{fig:render_speed_vs_isaac}
\end{figure*}

\textbf{Stability and Scalability in Complex Scenes.}
\label{sec:complexity_scene}
We evaluate the algorithmic robustness of \method against MuJoCo (CPU) and Genesis/MjWarp (GPU) in high-complexity single-environment scenarios. To vary scene complexity, we scale the number ($N$) of 27-DoF humanoid agents within a single environment. 
All experiments were conducted on an AMD 9950x CPU and an NVIDIA RTX 5090 GPU.
Results are shown in Fig. \ref{fig:benchmark}.
As constraint density increases, GPU-based solvers suffers from severe performance degradation. 
At $N=10$, Genesis fails to reach convergence and exhibits numerical instability through Jacobian-related errors. When complexity reaches $N=50$, MjWarp’s throughput collapses to a mere 1.71 FPS.
In contrast, \method (CPU) maintains a robust throughput of 1,015 FPS at $N=50$. This performance represents a 32$\times$ speedup over MuJoCo and a $\sim$600$\times$ improvement over MjWarp. 
Note that \method also offers competitive GPU performance, with further gains expected as we are refining our GPU-specific kernel fusion and memory management strategies.

\begin{table}[t]
    \centering
    \caption{\small \textbf{Image Quality Metrics.} Our pruned model retains 30\% of the original Gaussians while preserving high PSNR and SSIM for static scene reconstruction. }

    \scalebox{1}{
    \begin{tabular}{lcccc}
        \toprule
        \textbf{Method} 
        & \textbf{\# Gaussians $\downarrow$} 
        & \textbf{PSNR $\uparrow$} 
        & \textbf{SSIM $\uparrow$}
        & \textbf{LPIPS $\downarrow$} \\
        \midrule
        3DGS & 100\% & 27.1503 & 0.8296 & 0.2238 \\
        Ours            & 30\%  & 26.8658 & 0.8022 & 0.2840 \\
        \bottomrule
    \end{tabular}
    }
    \label{tab:visual_loss_com}
\end{table}

\begin{figure}[t]
  \centering
  \includegraphics[width=\linewidth]{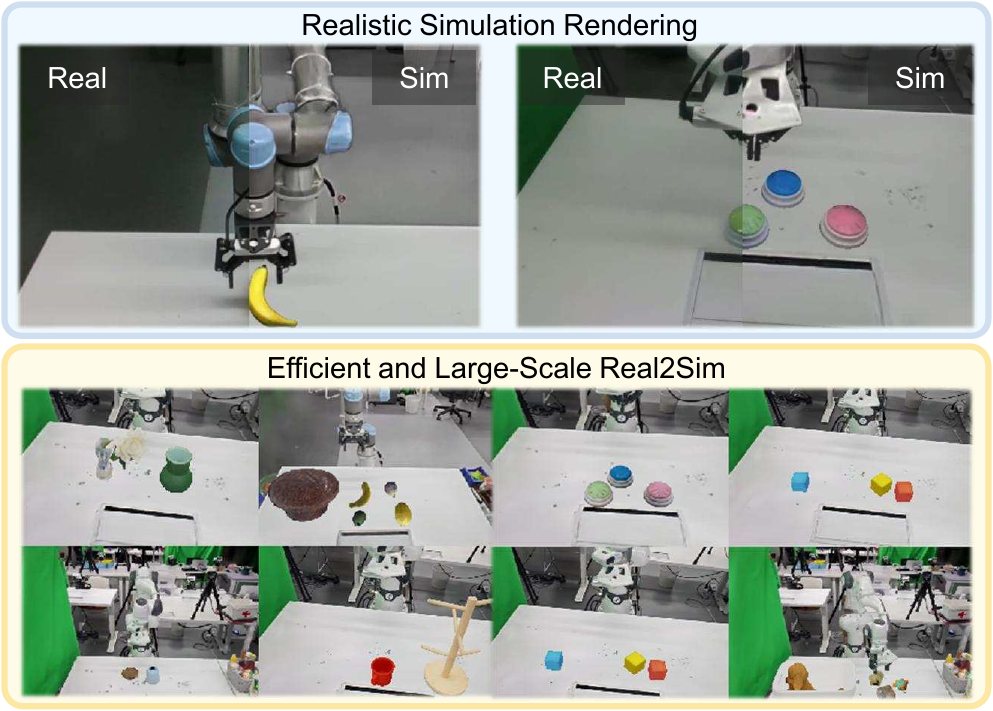}
  \vspace{-4mm}
  \caption{\textbf{Visualization of rendering}. The simulated renderings are nearly indistinguishable from real photographs 
  indicating photorealistic fidelity. 
  Our framework supports a broad range of tasks and evaluations, including diverse objects and scene configurations.
  }
  \vspace{-5mm}
  \label{fig:rendering_visualization}
\end{figure}

\subsection{Visual Fidelity and Efficiency}

\textbf{Visual 3DGS Compression.}
We measure the trade-off between memory footprint and visual quality, as summarized in Table~\ref{tab:visual_loss_com}. 
For full static scene reconstruction, we compare raw 3DGS reconstructions with our pruned version, retaining only about 30\% of the Gaussians while preserving high PSNR and SSIM. 
Additionally, for manipulated dynamic objects and the robot body, the number of points can be further reduced by up to 90\%, enabling more aggressive memory compression without compromising the critical visual cues required for robotic perception.

\textbf{Rendering Throughput Comparison}.
We evaluate the rendering throughput of \method against Isaac Sim's ray-tracing renderer across three standard resolutions and multiple GPU architectures, including the NVIDIA RTX 4090, RTX 6000 Ada, and A100. As illustrated in Fig. \ref{fig:render_speed_vs_isaac}, our framework consistently outperforms the baseline in aggregate FPS at all tested batch sizes, with the gap most pronounced at higher resolutions such as 1280$\times$720, where Isaac Sim's ray-tracing approach frequently encounters Out-Of-Memory (OOM) exceptions at larger batch sizes. Our Gaussian Splatting-based pipeline thus provides a more memory-efficient and scalable solution for large-scale parallel visual simulation.

\textbf{Qualitative Rendering Fidelity and Diversity.}
Fig.~\ref{fig:rendering_visualization} presents a qualitative comparison between real and simulated renderings.
The simulated renders exhibit a high degree of visual consistency with real camera images, preserving critical geometric features and surface details.
Moreover, the scenes depicted span a diverse set of object types, materials, and configurations, demonstrating that our simulation system maintains a high level of visual consistency across varied setups,
making it suitable for diverse training and evaluation scenarios.
All experiments were conducted on the Bridge-v2 dataset~\cite{walke2023bridgedata}. On an NVIDIA RTX 3090 GPU, the per-scene pipeline latency is within 5 minutes end-to-end before 3DGS pruning, while dataset-scale generation distributes pipeline stages across multiple GPUs. Compared to Bridge-v2, Bridge-GS enriches each asset with scene- and object-level 3DGS representations, object-level meshes, object poses, and camera intrinsics and extrinsics; Sec.~\ref{app:bridge_gs} provides additional dataset details.

\subsection{Physics-Intensive Locomotion Learning}

\begin{figure}[t]
    \centering
    \includegraphics[width=1\linewidth]{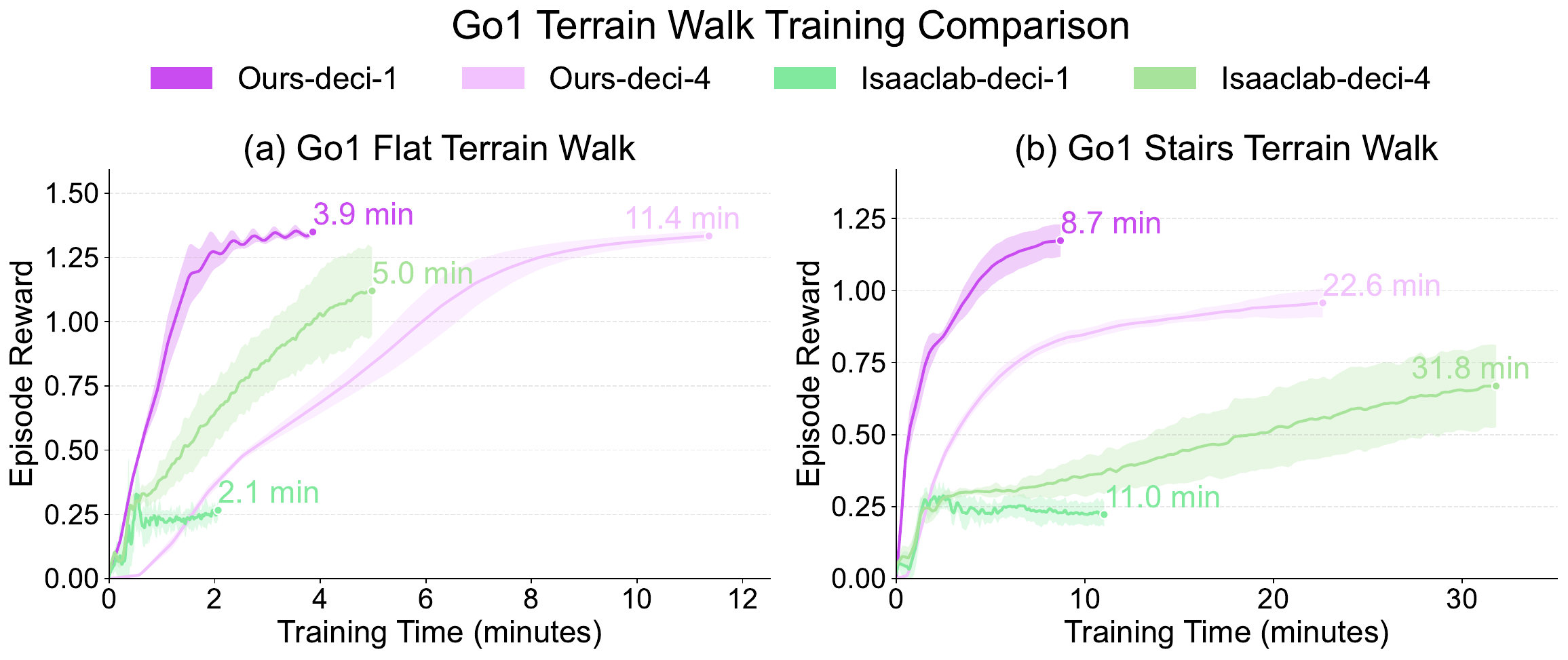}
    \caption{\textbf{Wall-clock training efficiency for Unitree Go1 locomotion.}
    (a) Flat terrain; (b) Rough terrain (stairs). ``deci'' denotes the decimation, which refers to the number of physical sub-steps per control step. Lower decimation typically increases throughput but may compromise physical fidelity. All results---both \method and the IsaacLab baseline---were measured on identical hardware (RTX 5070Ti + Ryzen 7 9700X).
    }
    \label{fig:go1_benchmark}
    \vspace{-5mm}
\end{figure}

\textbf{Simulation Comparison}.
We benchmark our framework against IsaacLab using the \textit{Isaac-Velocity-Flat-Unitree-Go1-v0} and \textit{Isaac-Velocity-Rough-Unitree-Go1-v0} environments. 
All configurations are trained for an equivalent number of total steps.
On flat terrain (Fig. \ref{fig:go1_benchmark}a), while IsaacLab achieves higher speed at low fidelity ($d=1$, where $d$ denotes the decimation factor, or the number of physical sub-steps per control step), it fails to reach a competitive terminal reward. In contrast, our method with $d=1$ achieves a terminal reward comparable to IsaacLab at $d=4$, while reaching convergence faster. This result demonstrates that our solver's stability permits larger integration time steps without compromising physical fidelity or policy convergence.
This stability advantage is even more apparent in complex environments like the stairs terrain (Fig. \ref{fig:go1_benchmark}b). Here, our framework at $d=1$ achieves higher rewards and faster convergence than the baseline at the same setting. Against the high-precision baseline ($d=4$), our method remains faster in wall-clock time. These results show that our approach offers a clear dual advantage in stability and speed for complex, contact-rich tasks.

\begin{figure}[t!]
\centering
\includegraphics[width=\linewidth]{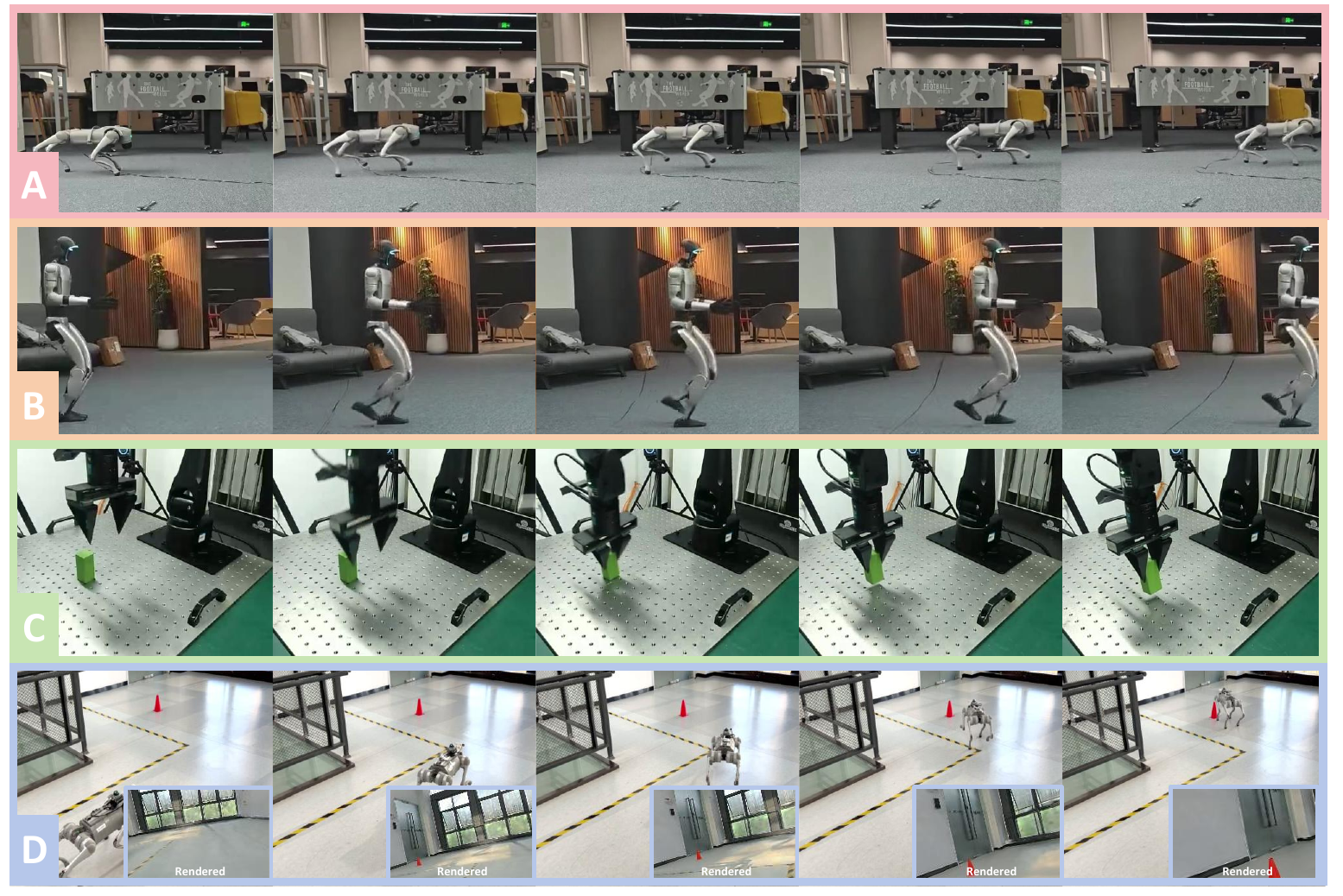}
\vspace{-4mm}
\caption{
\textbf{Real-world deployment of policies trained in GS-Playground}. We demonstrate robust Sim2Real transfer across diverse embodiments and modalities: \textbf{(a) Quadrupedal Locomotion}: Velocity tracking on Unitree Go2; \textbf{(b) Humanoid Locomotion}: 23-DoF balancing and walking on Unitree G1; \textbf{(c) Visual Manipulation}: End-to-end RGB-based grasping; \textbf{(d) Visual Navigation}: Real-time cone following on Unitree Go2 using raw RGB observations. 
}
\label{fig:tasks}
\vspace{-4mm}
\end{figure}

\textbf{Sim2Real Deployment}. 
We demonstrate the practical utility of \method by successfully deploying state-based locomotion policies onto a Unitree Go2 quadruped and a Unitree G1 humanoid (Fig. \ref{fig:tasks}(a), (b)). 
The quadruped policy, utilizing simplified collision geometries and 1,024 parallel environments, reached convergence in 10 minutes of wall-clock time. The humanoid policy, employing full-collision manifolds and 2,048 parallel environments, reached convergence in approximately 6 hours.
These results demonstrate our framework's efficiency in bridging the physics reality gap.

\subsection{Vision-Centric Navigation Learning}

We demonstrate a vision-based navigation task on the Unitree Go2, where the robot is required to search for and reach a target traffic cone (Fig.~\ref{fig:tasks}(d)).
The policy observes egocentric RGB images rendered by \method during training.
To couple high-level visual decision making with stable legged control, we adopt a two-level hierarchical RL design.
A \textit{high-level} policy encodes egocentric RGB observations and outputs a compact navigation command (e.g., desired base motion command).
A \textit{low-level} locomotion controller takes the command together with proprioceptive states and produces joint-level control signals for the Go2.
Both levels use PPO: the low-level controller is pre-trained and frozen, while the high-level policy is trained in \method.
After training in simulation, we directly deploy the learned policy on a real Go2, which successfully performs goal-directed navigation toward the cone using only onboard egocentric vision.
These results validate that \method provides the high-fidelity visual feedback necessary for training vision-encoder policies capable of zero-shot real-world deployment.

\subsection{Vision-Centric Manipulation Learning}
\label{subsec:manipulation}

\begin{figure}[t]
    \centering
    \includegraphics[width=\linewidth]{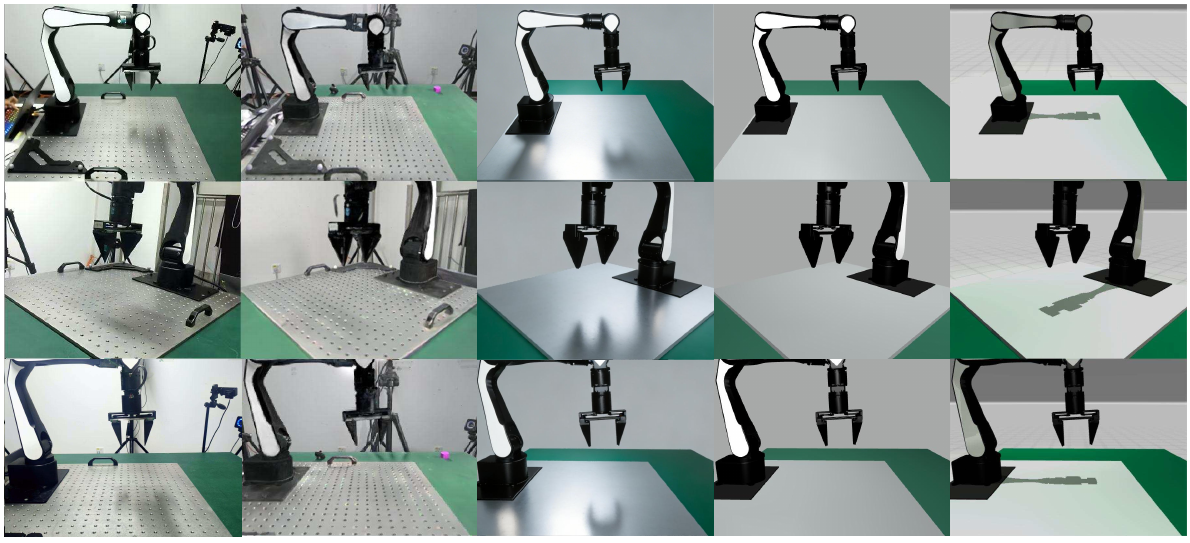}
    \vspace{-4mm}
    \caption{\textbf{Rendering comparison for the AIRBOT PickCube scene.} From left to right: Real World, \method, Isaac Lab, ManiSkill3, and Mujoco Playground. \method closely matches the appearance of the real scene, while all other simulators exhibit a substantial visual domain gap that, under the matched setup of Table~\ref{tab:sim_comparison}, leads to 0\% Sim2Real transfer.}
    \vspace{-3mm}
    \label{fig:render_comparison}
\end{figure}

To evaluate the platform's efficacy in bridging the visual Sim2Real gap, we conducted a block-grasping task using the Airbot Play robotic arm (Fig. \ref{fig:tasks}c). The control policy, trained as a goal-conditioned RL agent, maps raw RGB observations and proprioceptive states to 6-DoF joint actions. Utilizing our Real2Sim pipeline, we reconstructed a high-fidelity 3DGS simulation scene that serves as a direct visual proxy for the real-world setup. To improve robustness against real-world variability, we incorporated domain randomization of camera poses and image-level appearance (brightness, contrast, and exposure) during training. For comparison, we trained the same policy under matched task definition, network, PPO hyperparameters, and randomization protocol in three batched-rendering baselines---Mujoco Playground, ManiSkill3, and Isaac Lab---while varying the simulator and visual rendering stack.
The resulting policy achieves an 18/20 success rate (90\%) during zero-shot real-world deployment, whereas all three baselines achieve 0/20 under the same real-world evaluation protocol (Table~\ref{tab:sim_comparison} and Fig.~\ref{fig:render_comparison}). Under this matched setup, the observed failures are consistent with a visual domain gap between the baseline renderings and the real scene, rather than a difference in task definition or policy architecture. Notably, the evaluation was performed in a real-world scene that featured no specialized visual engineering or simplification, such as simplified backgrounds or controlled lighting. The robot demonstrates agile and stable grasping maneuvers, indicating that the perceptual richness of 3D-Gaussian-based simulation can support policies trained from complex, unsimplified visual cues. Sec.~\ref{app:manipulation} provides additional implementation details.

\begin{table}[t]
    \centering
    \caption{\textbf{Zero-shot Sim2Real success rate on PickCube} under matched task setup, network, and image-level domain randomization (20 real-world trials each).}
    \label{tab:sim_comparison}
    \scalebox{1}{
    \begin{tabular}{lc}
        \toprule
        \textbf{Simulator} & \textbf{Real-world Success Rate} \\
        \midrule
        Mujoco Playground & 0\,\%\,(0/20) \\
        ManiSkill3 & 0\,\%\,(0/20) \\
        Isaac Lab & 0\,\%\,(0/20) \\
        \textbf{\method (Ours)} & \textbf{90\,\%\,(18/20)} \\
        \bottomrule
    \end{tabular}
    }
    \vspace{-3mm}
\end{table}

\section{Discussion}
\label{sec:limitations}

We plan to utilize \method to synthesize large-scale vision-informed data for VLA and VLN models, facilitating robust sim-to-real transfer. Additionally, by incorporating the real-to-sim workflows, we are constructing expansive, scalable environments for the rigorous verification and benchmarking of advanced robotic policies.

Despite its current performance, our framework has several limitations that provide opportunities for future research. \method currently focuses on interactive rigid-body robot learning across manipulation, locomotion, and navigation. More general Real2Sim, including automatic estimation of physical parameters from visual observations alone, remains future work.

Unlike ray tracing or standard rasterization-based renderers, 3D Gaussian Splatting bakes static lighting into the reconstructed scene. Visual augmentation such as color jitter and brightness randomization improves robustness, but does not fully solve dynamic lighting. Transparent or highly reflective objects can also remain difficult even with segmentation and inpainting, especially under severe transparency, specular reflections, or heavy occlusion.

Furthermore, the current Rigid-Link Gaussian Kinematics (RLGK) assumes rigid bodies. Representing deformable objects (cloth, fluids) or soft-body manipulation remains a challenge. We plan to integrate particle-based dynamics (like PBD or MPIM) with Gaussian splatting to address non-rigid interactions.

\section{Conclusion}
\label{sec:conclusion}

We introduce \method, a high-performance simulation platform that harmonizes a custom parallel physics engine with a memory-efficient batch 3D Gaussian Splatting (3DGS) renderer to bridge the gap between realism and efficiency. By utilizing a specialized point-pruning strategy, our framework achieves an aggregate rendering throughput of $10^4$ FPS across 2048 environments at $640\times480$ resolution while avoiding the heavy memory overhead of traditional neural rendering. An automated Real2Sim pipeline streamlines the creation of simulation-ready visual and geometric assets for complex robot-learning scenes. Evaluations across quadrupedal, humanoid, and manipulation embodiments demonstrate that \method facilitates robust Sim2Real transfer by bridging the reality gap in both physical dynamics and visual perception.
Together, these capabilities establish \method as a scalable, open-source platform for high-throughput photorealistic robot learning and Sim2Real research.

\section{Acknowledgments}
\label{sec:acknowledgments}

The authors gratefully acknowledge support from D-Robotics under Grant No. 20243000104. We thank the MuJoCo and MuJoCo Playground teams for making their codebases available to the community; their software and documentation provided valuable references and infrastructure support for this work.

\bibliographystyle{plainnat}
\bibliography{references}

@article{kim2024openvla,
  title={Openvla: An open-source vision-language-action model},
  author={Kim, Moo Jin and Pertsch, Karl and Karamcheti, Siddharth and Xiao, Ted and Balakrishna, Ashwin and Nair, Suraj and Rafailov, Rafael and Foster, Ethan and Lam, Grace and Sanketi, Pannag and others},
  journal={arXiv preprint arXiv:2406.09246},
  year={2024}
}

@article{black2024pi_0,
  title={$\pi_0$: A Vision-Language-Action Flow Model for General Robot Control},
  author={Black, Kevin and Brown, Noah and Driess, Danny and Esmail, Adnan and Equi, Michael and Finn, Chelsea and Fusai, Niccolo and Groom, Lachy and Hausman, Karol and Ichter, Brian and others},
  journal={arXiv preprint arXiv:2410.24164},
  year={2024}
}

@article{zhou2025vision,
  title={Vision-Language-Action Model with Open-World Embodied Reasoning from Pretrained Knowledge},
  author={Zhou, Zhongyi and Zhu, Yichen and Wen, Junjie and Shen, Chaomin and Xu, Yi},
  journal={arXiv preprint arXiv:2505.21906},
  year={2025}
}

@article{hwangbo2019learning,
  title={Learning agile and dynamic motor skills for legged robots},
  author={Hwangbo, Jemin and Lee, Joonho and Dosovitskiy, Alexey and Bellicoso, Dario and Tsounis, Vassilios and Koltun, Vladlen and Hutter, Marco},
  journal={Science Robotics},
  volume={4},
  number={26},
  pages={eaau5872},
  year={2019},
  publisher={American Association for the Advancement of Science}
}

@article{kumar2021rma,
  title={Rma: Rapid motor adaptation for legged robots},
  author={Kumar, Ashish and Fu, Zipeng and Pathak, Deepak and Malik, Jitendra},
  journal={arXiv preprint arXiv:2107.04034},
  year={2021}
}

@article{siekmann2021blind,
  title={Blind bipedal stair traversal via sim-to-real reinforcement learning},
  author={Siekmann, Jonah and Green, Kevin and Warila, John and Fern, Alan and Hurst, Jonathan},
  journal={arXiv preprint arXiv:2105.08328},
  year={2021}
}

@inproceedings{margolis2023walk,
  title={Walk these ways: Tuning robot control for generalization with multiplicity of behavior},
  author={Margolis, Gabriel B and Agrawal, Pulkit},
  booktitle={Conference on Robot Learning},
  pages={22--31},
  year={2023},
  organization={PMLR}
}

@article{choi2023learning,
  title={Learning quadrupedal locomotion on deformable terrain},
  author={Choi, Suyoung and Ji, Gwanghyeon and Park, Jeongsoo and Kim, Hyeongjun and Mun, Juhyeok and Lee, Jeong Hyun and Hwangbo, Jemin},
  journal={Science Robotics},
  volume={8},
  number={74},
  pages={eade2256},
  year={2023},
  publisher={American Association for the Advancement of Science}
}

@article{zhuang2023robot,
  title={Robot parkour learning},
  author={Zhuang, Ziwen and Fu, Zipeng and Wang, Jianren and Atkeson, Christopher and Schwertfeger, Soeren and Finn, Chelsea and Zhao, Hang},
  journal={arXiv preprint arXiv:2309.05665},
  year={2023}
}

@article{margolis2024rapid,
  title={Rapid locomotion via reinforcement learning},
  author={Margolis, Gabriel B and Yang, Ge and Paigwar, Kartik and Chen, Tao and Agrawal, Pulkit},
  journal={The International Journal of Robotics Research},
  volume={43},
  number={4},
  pages={572--587},
  year={2024},
  publisher={SAGE Publications Sage UK: London, England}
}

@inproceedings{wang2024arm,
  title={Arm-Constrained Curriculum Learning for Loco-Manipulation of a Wheel-Legged Robot},
  author={Wang, Zifan and Jia, Yufei and Shi, Lu and Wang, Haoyu and Zhao, Haizhou and Li, Xueyang and Zhou, Jinni and Ma, Jun and Zhou, Guyue},
  booktitle={2024 IEEE/RSJ International Conference on Intelligent Robots and Systems (IROS)},
  pages={10770--10776},
  year={2024},
  organization={IEEE}
}

@article{he2025asap,
  title={Asap: Aligning simulation and real-world physics for learning agile humanoid whole-body skills},
  author={He, Tairan and Gao, Jiawei and Xiao, Wenli and Zhang, Yuanhang and Wang, Zi and Wang, Jiashun and Luo, Zhengyi and He, Guanqi and Sobanbab, Nikhil and Pan, Chaoyi and others},
  journal={arXiv preprint arXiv:2502.01143},
  year={2025}
}

@article{wang2025omni,
  title={Omni-Perception: Omnidirectional Collision Avoidance for Legged Locomotion in Dynamic Environments},
  author={Wang, Zifan and Ma, Teli and Jia, Yufei and Yang, Xun and Zhou, Jiaming and Ouyang, Wenlong and Zhang, Qiang and Liang, Junwei},
  journal={arXiv preprint arXiv:2505.19214},
  year={2025}
}

@article{shacklett2023extensible,
  title={An extensible, data-oriented architecture for high-performance, many-world simulation},
  author={Shacklett, Brennan and Rosenzweig, Luc Guy and Xie, Zhiqiang and Sarkar, Bidipta and Szot, Andrew and Wijmans, Erik and Koltun, Vladlen and Batra, Dhruv and Fatahalian, Kayvon},
  journal={ACM Transactions on Graphics (TOG)},
  volume={42},
  number={4},
  pages={1--13},
  year={2023},
  publisher={ACM New York, NY, USA}
}

@inproceedings{todorov2012mujoco,
  title={Mujoco: A physics engine for model-based control},
  author={Todorov, Emanuel and Erez, Tom and Tassa, Yuval},
  booktitle={2012 IEEE/RSJ international conference on intelligent robots and systems},
  pages={5026--5033},
  year={2012},
  organization={IEEE}
}

@article{zakka2025mujoco,
  title={Mujoco playground},
  author={Zakka, Kevin and Tabanpour, Baruch and Liao, Qiayuan and Haiderbhai, Mustafa and Holt, Samuel and Luo, Jing Yuan and Allshire, Arthur and Frey, Erik and Sreenath, Koushil and Kahrs, Lueder A and others},
  journal={arXiv preprint arXiv:2502.08844},
  year={2025}
}

@article{makoviychuk2021isaac,
  title={Isaac gym: High performance gpu-based physics simulation for robot learning},
  author={Makoviychuk, Viktor and Wawrzyniak, Lukasz and Guo, Yunrong and Lu, Michelle and Storey, Kier and Macklin, Miles and Hoeller, David and Rudin, Nikita and Allshire, Arthur and Handa, Ankur and others},
  journal={arXiv preprint arXiv:2108.10470},
  year={2021}
}

@article{mittal2025isaac,
  title={Isaac lab: A gpu-accelerated simulation framework for multi-modal robot learning},
  author={Mittal, Mayank and Roth, Pascal and Tigue, James and Richard, Antoine and Zhang, Octi and Du, Peter and Serrano-Mu{\~n}oz, Antonio and Yao, Xinjie and Zurbr{\"u}gg, Ren{\'e} and Rudin, Nikita and others},
  journal={arXiv preprint arXiv:2511.04831},
  year={2025}
}

@article{taomaniskill3,
  title={ManiSkill3: GPU Parallelized Robotics Simulation and Rendering for Generalizable Embodied AI},
  author={Stone Tao and Fanbo Xiang and Arth Shukla and Yuzhe Qin and Xander Hinrichsen and Xiaodi Yuan and Chen Bao and Xinsong Lin and Yulin Liu and Tse-kai Chan and Yuan Gao and Xuanlin Li and Tongzhou Mu and Nan Xiao and Arnav Gurha and Viswesh Nagaswamy Rajesh and Yong Woo Choi and Yen-Ru Chen and Zhiao Huang and Roberto Calandra and Rui Chen and Shan Luo and Hao Su},
  journal = {Robotics: Science and Systems},
  year={2025},
}

@article{Genesis,
  title={Genesis: A Generative and Universal Physics Engine for Robotics and Beyond},
  author={Zhou, Xian and Qiao, Yiling and Xu, Zhenjia and Wang, TH and Chen, Z and Zheng, J and Xiong, Z and Wang, Y and Zhang, M and Ma, P and others},
  journal={arXiv preprint arXiv:2401.01454},
  year={2024}
}

@article{akkaya2019solving,
  title={Solving rubik's cube with a robot hand},
  author={Akkaya, Ilge and Andrychowicz, Marcin and Chociej, Maciek and Litwin, Mateusz and McGrew, Bob and Petron, Arthur and Paino, Alex and Plappert, Matthias and Powell, Glenn and Ribas, Raphael and others},
  journal={arXiv preprint arXiv:1910.07113},
  year={2019}
}

@article{zhai2025vision,
  title={A vision-language-action-critic model for robotic real-world reinforcement learning},
  author={Zhai, Shaopeng and Zhang, Qi and Zhang, Tianyi and Huang, Fuxian and Zhang, Haoran and Zhou, Ming and Zhang, Shengzhe and Liu, Litao and Lin, Sixu and Pang, Jiangmiao},
  journal={arXiv preprint arXiv:2509.15937},
  year={2025}
}

@article{li2025simplevla,
  title={Simplevla-rl: Scaling vla training via reinforcement learning},
  author={Li, Haozhan and Zuo, Yuxin and Yu, Jiale and Zhang, Yuhao and Yang, Zhaohui and Zhang, Kaiyan and Zhu, Xuekai and Zhang, Yuchen and Chen, Tianxing and Cui, Ganqu and others},
  journal={arXiv preprint arXiv:2509.09674},
  year={2025}
}

@article{xue2025opening,
  title={Opening the Sim-to-Real Door for Humanoid Pixel-to-Action Policy Transfer},
  author={Xue, Haoru and He, Tairan and Wang, Zi and Ben, Qingwei and Xiao, Wenli and Luo, Zhengyi and Da, Xingye and Casta{\~n}eda, Fernando and Shi, Guanya and Sastry, Shankar and others},
  journal={arXiv preprint arXiv:2512.01061},
  year={2025}
}

@article{he2025viral,
  title={VIRAL: Visual Sim-to-Real at Scale for Humanoid Loco-Manipulation},
  author={He, Tairan and Wang, Zi and Xue, Haoru and Ben, Qingwei and Luo, Zhengyi and Xiao, Wenli and Yuan, Ye and Da, Xingye and Casta{\~n}eda, Fernando and Sastry, Shankar and others},
  journal={arXiv preprint arXiv:2511.15200},
  year={2025}
}

@article{cao2025learning,
  title={Learning Motion Skills with Adaptive Assistive Curriculum Force in Humanoid Robots},
  author={Cao, Zhanxiang and Zhang, Yang and Nie, Buqing and Lin, Huangxuan and Li, Haoyang and Gao, Yue},
  journal={arXiv preprint arXiv:2506.23125},
  year={2025}
}

@article{zhang2025keep,
  title={Keep on Going: Learning Robust Humanoid Motion Skills via Selective Adversarial Training},
  author={Zhang, Yang and Cao, Zhanxiang and Nie, Buqing and Li, Haoyang and Jiangwei, Zhong and Sun, Qiao and Hu, Xiaoyi and Yang, Xiaokang and Gao, Yue},
  journal={arXiv preprint arXiv:2507.08303},
  year={2025}
}

@inproceedings{cheng2024navila,
title = {NaVILA: Legged Robot Vision-Language-Action Model for Navigation},
    author = {Cheng, An-Chieh and Ji, Yandong and Yang, Zhaojing and Zou, Xueyan and Kautz, Jan and Biyik, Erdem and Yin,
    Hongxu and Liu, Sifei and Wang, Xiaolong},
    booktitle = {RSS},
    year = {2025},
}

@article{cai2025navdp,
  title={{NavDP}: Learning Sim-to-Real Navigation Diffusion Policy with Privileged Information Guidance},
  author={Cai, Wenzhe and Peng, Jiaqi and Yang, Yuqiang and Zhang, Yujian and Wei, Meng and Wang, Hanqing and Chen, Yilun and Wang, Tai and Pang, Jiangmiao},
  journal={arXiv preprint arXiv:2501.04610},
  year={2025}
}

@article{zhang2024uninavid,
  title={Uni-navid: A video-based vision-language-action model for unifying embodied navigation tasks},
  author={Zhang, Jiazhao and Wang, Kunyu and Wang, Shaoan and Li, Minghan and Liu, Haoran and Wei, Songlin and Wang, Zhongyuan and Zhang, Zhizheng and Wang, He},
  journal={arXiv preprint arXiv:2412.06224},
  year={2024}
}

@article{zhang2025embodied,
  title={Embodied navigation foundation model},
  author={Zhang, Jiazhao and Li, Anqi and Qi, Yunpeng and Li, Minghan and Liu, Jiahang and Wang, Shaoan and Liu, Haoran and Zhou, Gengze and Wu, Yuze and Li, Xingxing and others},
  journal={arXiv preprint arXiv:2509.12129},
  year={2025}
}

@article{kerbl20233d,
  title={3D Gaussian splatting for real-time radiance field rendering.},
  author={Kerbl, Bernhard and Kopanas, Georgios and Leimk{\"u}hler, Thomas and Drettakis, George},
  journal={ACM Trans. Graph.},
  volume={42},
  number={4},
  pages={139--1},
  year={2023}
}

@article{ye2025gsplat,
  title={gsplat: An open-source library for Gaussian splatting},
  author={Ye, Vickie and Li, Ruilong and Kerr, Justin and Turkulainen, Matias and Yi, Brent and Pan, Zhuoyang and Seiskari, Otto and Ye, Jianbo and Hu, Jeffrey and Tancik, Matthew and Angjoo Kanazawa},
  journal={Journal of Machine Learning Research},
  volume={26},
  number={34},
  pages={1--17},
  year={2025}
}

@inproceedings{wang2025vggt,
  title={Vggt: Visual geometry grounded transformer},
  author={Wang, Jianyuan and Chen, Minghao and Karaev, Nikita and Vedaldi, Andrea and Rupprecht, Christian and Novotny, David},
  booktitle={Proceedings of the Computer Vision and Pattern Recognition Conference},
  pages={5294--5306},
  year={2025}
}

@article{mescheder2025sharp,
  title={Sharp Monocular View Synthesis in Less Than a Second},
  author={Mescheder, Lars and Dong, Wei and Li, Shiwei and Bai, Xuyang and Santos, Marcel and Hu, Peiyun and Lecouat, Bruno and Zhen, Mingmin and Delaunoy, Ama{\~A}{\c{G}}l and Fang, Tian and others},
  journal={arXiv preprint arXiv:2512.10685},
  year={2025}
}

@article{jiang2025anysplat,
  title={Anysplat: Feed-forward 3d gaussian splatting from unconstrained views},
  author={Jiang, Lihan and Mao, Yucheng and Xu, Linning and Lu, Tao and Ren, Kerui and Jin, Yichen and Xu, Xudong and Yu, Mulin and Pang, Jiangmiao and Zhao, Feng and others},
  journal={ACM Transactions on Graphics (TOG)},
  volume={44},
  number={6},
  pages={1--16},
  year={2025},
  publisher={ACM New York, NY, USA}
}

@inproceedings{hanson2025pup,
  title={Pup 3d-gs: Principled uncertainty pruning for 3d gaussian splatting},
  author={Hanson, Alex and Tu, Allen and Singla, Vasu and Jayawardhana, Mayuka and Zwicker, Matthias and Goldstein, Tom},
  booktitle={Proceedings of the Computer Vision and Pattern Recognition Conference},
  pages={5949--5958},
  year={2025}
}

@inproceedings{fang2024mini,
  title={Mini-splatting: Representing scenes with a constrained number of gaussians},
  author={Fang, Guangchi and Wang, Bing},
  booktitle={European Conference on Computer Vision},
  pages={165--181},
  year={2024},
  organization={Springer}
}

@inproceedings{hanson2025speedy,
  title={Speedy-splat: Fast 3d gaussian splatting with sparse pixels and sparse primitives},
  author={Hanson, Alex and Tu, Allen and Lin, Geng and Singla, Vasu and Zwicker, Matthias and Goldstein, Tom},
  booktitle={Proceedings of the Computer Vision and Pattern Recognition Conference},
  pages={21537--21546},
  year={2025}
}

@article{chen2025sam,
  title={Sam 3d: 3dfy anything in images},
  author={Chen, Xingyu and Chu, Fu-Jen and Gleize, Pierre and Liang, Kevin J and Sax, Alexander and Tang, Hao and Wang, Weiyao and Guo, Michelle and Hardin, Thibaut and Li, Xiang and others},
  journal={arXiv preprint arXiv:2511.16624},
  year={2025}
}

@article{lin2025diffsplat,
  title={Diffsplat: Repurposing image diffusion models for scalable gaussian splat generation},
  author={Lin, Chenguo and Pan, Panwang and Yang, Bangbang and Li, Zeming and Mu, Yadong},
  journal={arXiv preprint arXiv:2501.16764},
  year={2025}
}

@inproceedings{xia2024video2game,
  title={Video2game: Real-time interactive realistic and browser-compatible environment from a single video},
  author={Xia, Hongchi and Lin, Zhi-Hao and Ma, Wei-Chiu and Wang, Shenlong},
  booktitle={Proceedings of the IEEE/CVF Conference on Computer Vision and Pattern Recognition},
  pages={4578--4588},
  year={2024}
}

@inproceedings{xie2025vid2sim,
  title={Vid2sim: Realistic and interactive simulation from video for urban navigation},
  author={Xie, Ziyang and Liu, Zhizheng and Peng, Zhenghao and Wu, Wayne and Zhou, Bolei},
  booktitle={Proceedings of the Computer Vision and Pattern Recognition Conference},
  pages={1581--1591},
  year={2025}
}

@article{li2024robogsim,
  title={Robogsim: A real2sim2real robotic gaussian splatting simulator},
  author={Li, Xinhai and Li, Jialin and Zhang, Ziheng and Zhang, Rui and Jia, Fan and Wang, Tiancai and Fan, Haoqiang and Tseng, Kuo-Kun and Wang, Ruiping},
  journal={arXiv preprint arXiv:2411.11839},
  year={2024}
}

@article{han2025re,
  title={Re$^3$ Sim: Generating High-Fidelity Simulation Data via 3D-Photorealistic Real-to-Sim for Robotic Manipulation},
  author={Han, Xiaoshen and Liu, Minghuan and Chen, Yilun and Yu, Junqiu and Lyu, Xiaoyang and Tian, Yang and Wang, Bolun and Zhang, Weinan and Pang, Jiangmiao},
  journal={arXiv preprint arXiv:2502.08645},
  year={2025}
}

@inproceedings{qureshi2025splatsim,
  title={Splatsim: Zero-shot sim2real transfer of rgb manipulation policies using gaussian splatting},
  author={Qureshi, M Nomaan and Garg, Sparsh and Yandun, Francisco and Held, David and Kantor, George and Silwal, Abhisesh},
  booktitle={2025 IEEE International Conference on Robotics and Automation (ICRA)},
  pages={6502--6509},
  year={2025},
  organization={IEEE}
}

@article{jia2025discoverse,
  title={Discoverse: Efficient robot simulation in complex high-fidelity environments},
  author={Jia, Yufei and Wang, Guangyu and Dong, Yuhang and Wu, Junzhe and Zeng, Yupei and Lin, Haonan and Wang, Zifan and Ge, Haizhou and Gu, Weibin and Ding, Kairui and others},
  journal={arXiv preprint arXiv:2507.21981},
  year={2025}
}

@article{escontrela2025gaussgym,
  title={GaussGym: An open-source real-to-sim framework for learning locomotion from pixels},
  author={Escontrela, Alejandro and Kerr, Justin and Allshire, Arthur and Frey, Jonas and Duan, Rocky and Sferrazza, Carmelo and Abbeel, Pieter},
  journal={arXiv preprint arXiv:2510.15352},
  year={2025}
}

@article{barcellona2024dream,
  title={Dream to manipulate: Compositional world models empowering robot imitation learning with imagination},
  author={Barcellona, Leonardo and Zadaianchuk, Andrii and Allegro, Davide and Papa, Samuele and Ghidoni, Stefano and Gavves, Efstratios},
  journal={arXiv preprint arXiv:2412.14957},
  year={2024}
}

@inproceedings{lou2025robo,
  title={Robo-gs: A physics consistent spatial-temporal model for robotic arm with hybrid representation},
  author={Lou, Haozhe and Liu, Yurong and Pan, Yike and Geng, Yiran and Chen, Jianteng and Ma, Wenlong and Li, Chenglong and Wang, Lin and Feng, Hengzhen and Shi, Lu and others},
  booktitle={2025 IEEE International Conference on Robotics and Automation (ICRA)},
  pages={15379--15386},
  year={2025},
  organization={IEEE}
}

@article{yu2025real2render2real,
  title={Real2render2real: Scaling robot data without dynamics simulation or robot hardware},
  author={Yu, Justin and Fu, Letian and Huang, Huang and El-Refai, Karim and Ambrus, Rares Andrei and Cheng, Richard and Irshad, Muhammad Zubair and Goldberg, Ken},
  journal={arXiv preprint arXiv:2505.09601},
  year={2025}
}

@article{yang2025novel,
  title={Novel demonstration generation with gaussian splatting enables robust one-shot manipulation},
  author={Yang, Sizhe and Yu, Wenye and Zeng, Jia and Lv, Jun and Ren, Kerui and Lu, Cewu and Lin, Dahua and Pang, Jiangmiao},
  journal={arXiv preprint arXiv:2504.13175},
  year={2025}
}

@article{jiang2025phystwin,
  title={Phystwin: Physics-informed reconstruction and simulation of deformable objects from videos},
  author={Jiang, Hanxiao and Hsu, Hao-Yu and Zhang, Kaifeng and Yu, Hsin-Ni and Wang, Shenlong and Li, Yunzhu},
  journal={arXiv preprint arXiv:2503.17973},
  year={2025}
}

@article{abou2025real,
  title={Real-is-Sim: Bridging the Sim-to-Real Gap with a Dynamic Digital Twin for Real-World Robot Policy Evaluation},
  author={Abou-Chakra, Jad and Sun, Lingfeng and Rana, Krishan and May, Brandon and Schmeckpeper, Karl and Minniti, Maria Vittoria and Herlant, Laura},
  journal={arXiv preprint arXiv:2504.03597},
  year={2025}
}

@article{jiang2025gsworld,
  title={Gsworld: Closed-loop photo-realistic simulation suite for robotic manipulation},
  author={Jiang, Guangqi and Chang, Haoran and Qiu, Ri-Zhao and Liang, Yutong and Ji, Mazeyu and Zhu, Jiyue and Dong, Zhao and Zou, Xueyan and Wang, Xiaolong},
  journal={arXiv preprint arXiv:2510.20813},
  year={2025}
}

@article{zhang2025real,
  title={Real-to-sim robot policy evaluation with gaussian splatting simulation of soft-body interactions},
  author={Zhang, Kaifeng and Sha, Shuo and Jiang, Hanxiao and Loper, Matthew and Song, Hyunjong and Cai, Guangyan and Xu, Zhuo and Hu, Xiaochen and Zheng, Changxi and Li, Yunzhu},
  journal={arXiv preprint arXiv:2511.04665},
  year={2025}
}

@article{zhu2020robosuite,
  title={robosuite: A modular simulation framework and benchmark for robot learning},
  author={Zhu, Yuke and Wong, Josiah and Mandlekar, Ajay and Mart{\'\i}n-Mart{\'\i}n, Roberto and Joshi, Abhishek and Nasiriany, Soroush and Zhu, Yifeng},
  journal={arXiv preprint arXiv:2009.12293},
  year={2020}
}

@inproceedings{li2025clone,
  title={{CLONE}: Closed-Loop Whole-Body Humanoid Teleoperation for Long-Horizon Tasks},
  author={Li, Yixuan and Lin, Yutang and Cui, Jieming and Liu, Tengyu and Liang, Wei and Zhu, Yixin and Huang, Siyuan},
  booktitle={9th Annual Conference on Robot Learning (CoRL)},
  year={2025}
}

@article{chen2023visual,
  title={Visual dexterity: In-hand reorientation of novel and complex object shapes},
  author={Chen, Tao and Tippur, Megha and Wu, Siyang and Kumar, Vikash and Adelson, Edward and Agrawal, Pulkit},
  journal={Science Robotics},
  volume={8},
  number={84},
  pages={eadc9244},
  year={2023},
  publisher={American Association for the Advancement of Science}
}

@article{jiang2025robust,
  title={Robust In-Hand Reorientation With Hierarchical RL-Based Motion Primitives and Model-Based Regrasping},
  author={Jiang, Yongpeng and Yu, Mingrui and Chen, Chen and Jia, Yongyi and Li, Xiang},
  journal={IEEE Robotics and Automation Practice},
  volume={1},
  pages={12--17},
  year={2025},
  publisher={IEEE}
}

@inproceedings{yin2025dexteritygen,
  title={{DexterityGen}: Foundation Controller for Unprecedented Dexterity},
  author={Yin, Zhao-Heng and Wang, Changhao and Pineda, Luis and Hogan, Francois and Bodduluri, Krishna and Sharma, Akash and Lancaster, Patrick and Prasad, Ishita and Kalakrishnan, Mrinal and Malik, Jitendra and others},
  booktitle={Proceedings of Robotics: Science and Systems (RSS)},
  year={2025}
}

@inproceedings{matas2018sim,
  title={Sim-to-Real Reinforcement Learning for Deformable Object Manipulation},
  author={Matas, Jan and James, Stephen and Davison, Andrew J.},
  booktitle={Conference on Robot Learning (CoRL)},
  pages={734--743},
  year={2018},
  organization={PMLR}
}

@article{liu2023grounding,
  title={Grounding dino: Marrying dino with grounded pre-training for open-set object detection},
  author={Liu, Shilong and Zeng, Zhaoyang and Ren, Tianhe and Li, Feng and Zhang, Hao and Yang, Jie and Li, Chunyuan and Yang, Jianwei and Su, Hang and Zhu, Jun and others},
  journal={arXiv preprint arXiv:2303.05499},
  year={2023}
}

@article{kirillov2023segany,
  title={Segment Anything},
  author={Kirillov, Alexander and Mintun, Eric and Ravi, Nikhila and Mao, Hanzi and Rolland, Chloe and Gustafson, Laura and Xiao, Tete and Whitehead, Spencer and Berg, Alexander C. and Lo, Wan-Yen and Doll{\'a}r, Piotr and Girshick, Ross},
  journal={arXiv:2304.02643},
  year={2023}
}

@article{ravi2024sam2,
  title={SAM 2: Segment Anything in Images and Videos},
  author={Ravi, Nikhila and Gabeur, Valentin and Hu, Yuan-Ting and Hu, Ronghang and Ryali, Chaitanya and Ma, Tengyu and Khedr, Haitham and R{\"a}dle, Roman and Rolland, Chloe and Gustafson, Laura and Mintun, Eric and Pan, Junting and Alwala, Kalyan Vasudev and Carion, Nicolas and Wu, Chao-Yuan and Girshick, Ross and Doll{\'a}r, Piotr and Feichtenhofer, Christoph},
  journal={arXiv preprint arXiv:2408.00714},
  url={https://arxiv.org/abs/2408.00714},
  year={2024}
}

@inproceedings{walke2023bridgedata,
    title={BridgeData V2: A Dataset for Robot Learning at Scale},
    author={Walke, Homer and Black, Kevin and Lee, Abraham and Kim, Moo Jin and Du, Max and Zheng, Chongyi and Zhao, Tony and Hansen-Estruch, Philippe and Vuong, Quan and He, Andre and Myers, Vivek and Fang, Kuan and Finn, Chelsea and Levine, Sergey},
    booktitle={Conference on Robot Learning (CoRL)},
    year={2023}
}

@article{suvorov2021resolution,
  title={Resolution-robust Large Mask Inpainting with Fourier Convolutions},
  author={Suvorov, Roman and Logacheva, Elizaveta and Mashikhin, Anton and Remizova, Anastasia and Ashukha, Arsenii and Silvestrov, Aleksei and Kong, Naejin and Goka, Harshith and Park, Kiwoong and Lempitsky, Victor},
  journal={arXiv preprint arXiv:2109.07161},
  year={2021}
}

\clearpage
\onecolumn
\input{Sections/8_Appendix}

\end{document}